\documentclass[review]{elsarticle}
\journal{Pattern Recognition}

\usepackage{graphicx, xcolor}

\usepackage{amsmath, amssymb, amsthm}
\usepackage{mathtools}
\usepackage{physics}    

\usepackage{float}      
\usepackage{stfloats}   
\usepackage{caption}
\usepackage{subcaption}

\usepackage{lineno}
\modulolinenumbers[5]

\usepackage{todonotes}
\usepackage{etoolbox}
\usepackage[algo2e, ruled, noend, commentsnumbered]{algorithm2e}
\usepackage{hyperref}

\DeclareMathOperator{\ADTW}{ADTW}
\DeclareMathOperator{\DTW}{DTW}
\DeclareMathOperator{\CDTW}{CDTW}
\DeclareMathOperator{\WDTW}{WDTW}
\DeclareMathOperator{\BEST}{BEST}

\DeclareMathOperator{\SQED}{SQED}

\DeclareMathOperator{\Dtrain}{\mathcal{D}}
\DeclareMathOperator{\Dtest}{\mathcal{T}}

\DeclareMathOperator{\cost}{\gamma}
\DeclareMathOperator{\reverse}{reverse}

\DeclareMathOperator{\similarity}{dist}

\DeclareMathOperator{\Ali}{\mathcal{A}}

\DeclareMathOperator{\alen}{\lambda}
\DeclareMathOperator{\domain}{\mathbb{D}}
\DeclareMathOperator{\R}{\mathbb{R}}
\DeclareMathOperator{\weight}{weight}
\DeclareMathOperator{\mean}{mean}

\begin{document}

\begin{frontmatter}
\title{Amercing: An Intuitive, Elegant and Effective Constraint for Dynamic Time Warping}

\author{Matthieu Herrmann\corref{1}}
\cortext[1]{Corresponding author}
\ead{matthieu.herrmann@monash.edu}

\author{Geoffrey I. Webb}
\ead{geoff.webb@monash.edu}

\address{
Department of Data Science and Artificial Intelligence and Monash Data Futures Institute,
Monash University,
Melbourne, VIC 3800, Australia}

\begin{abstract}
Dynamic Time Warping ($\DTW$),
and its \emph{constrained} ($\CDTW$) and \emph{weighted} ($\WDTW$) variants,
are time series distances with a wide range of applications.
They minimize the cost of non-linear alignments between series.
$\CDTW$ and $\WDTW$ have been introduced because $\DTW$ is too permissive in its alignments.
However,
$\CDTW$ uses a crude step function, allowing unconstrained flexibility within the window, and none beyond it.
$\WDTW$'s multiplicative weight is relative to the distances between aligned points
along a warped path, rather than being a direct function of the amount of warping that is introduced.
In this paper, we introduce \emph{Amerced Dynamic Time Warping} ($\ADTW$),
a new, intuitive, $\DTW$ variant that penalizes the act of warping by a fixed additive cost.
Like $\CDTW$ and $\WDTW$, $\ADTW$ constrains the amount of warping.
However, it avoids both abrupt discontinuities in the amount of warping allowed
and the limitations of a multiplicative penalty.
We formally introduce $\ADTW$, prove some of its properties, and discuss its parameterization.
We show on a simple example how it can be parameterized to achieve an intuitive outcome,
and demonstrate its usefulness on a standard time series classification benchmark.
We provide a demonstration application in C++ \cite{demoapp}.
\end{abstract}


\end{frontmatter}


\section{Introduction}
Dynamic Time Warping ($\DTW$) is a distance measure for time series.
First developed for speech recognition~\cite{sakoe1971, sakoe1978},
it has been adopted across a broad spectrum of applications including
gesture recognition~\cite{cheng2016image},
signature verification~\cite{OKAWA2021107699},
shape matching~\cite{yasseen2016shape},
road surface monitoring~\cite{singh2017smart},
neuroscience~\cite{cao2016real}
and medical diagnosis~\cite{varatharajan2018wearable}.  

\begin{figure}
    \centering
    \begin{subfigure}[b]{0.31\textwidth}
        \includegraphics[width=\textwidth]{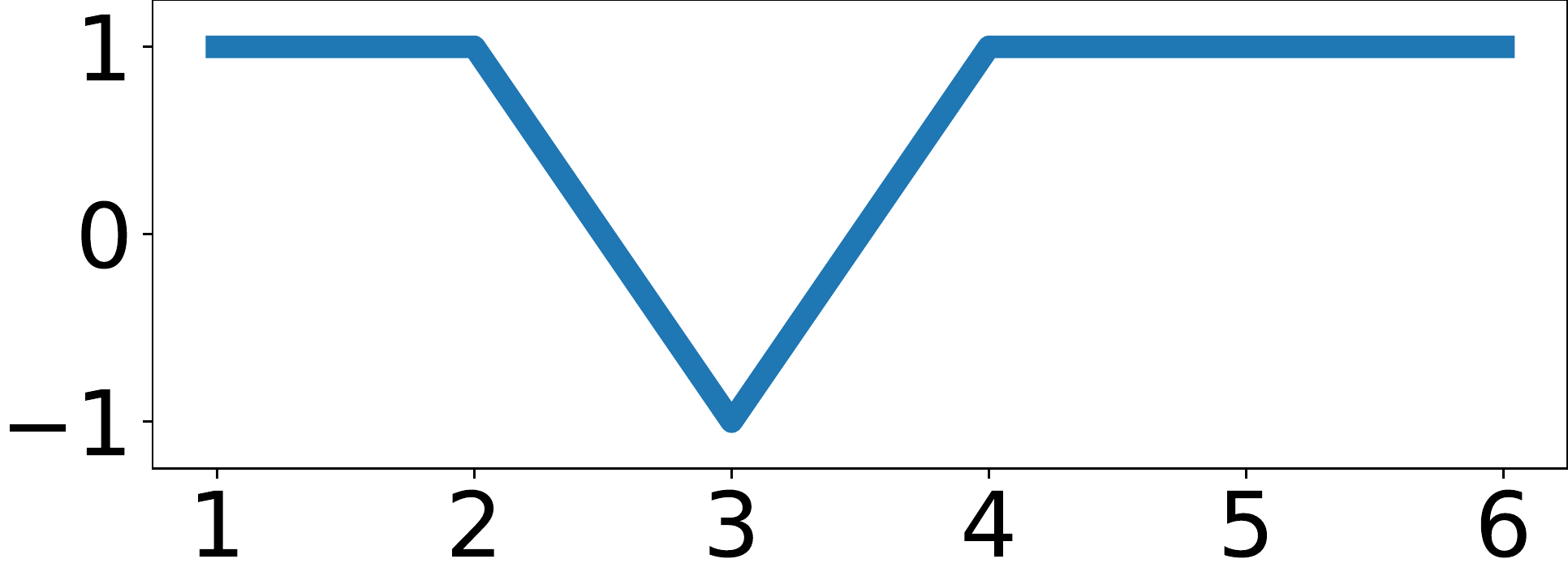}
        \caption{$S=\{1, 1, -1, 1, 1, 1\}$}
    \end{subfigure}
    \hfill
    \begin{subfigure}[b]{0.31\textwidth}
        \includegraphics[width=\textwidth]{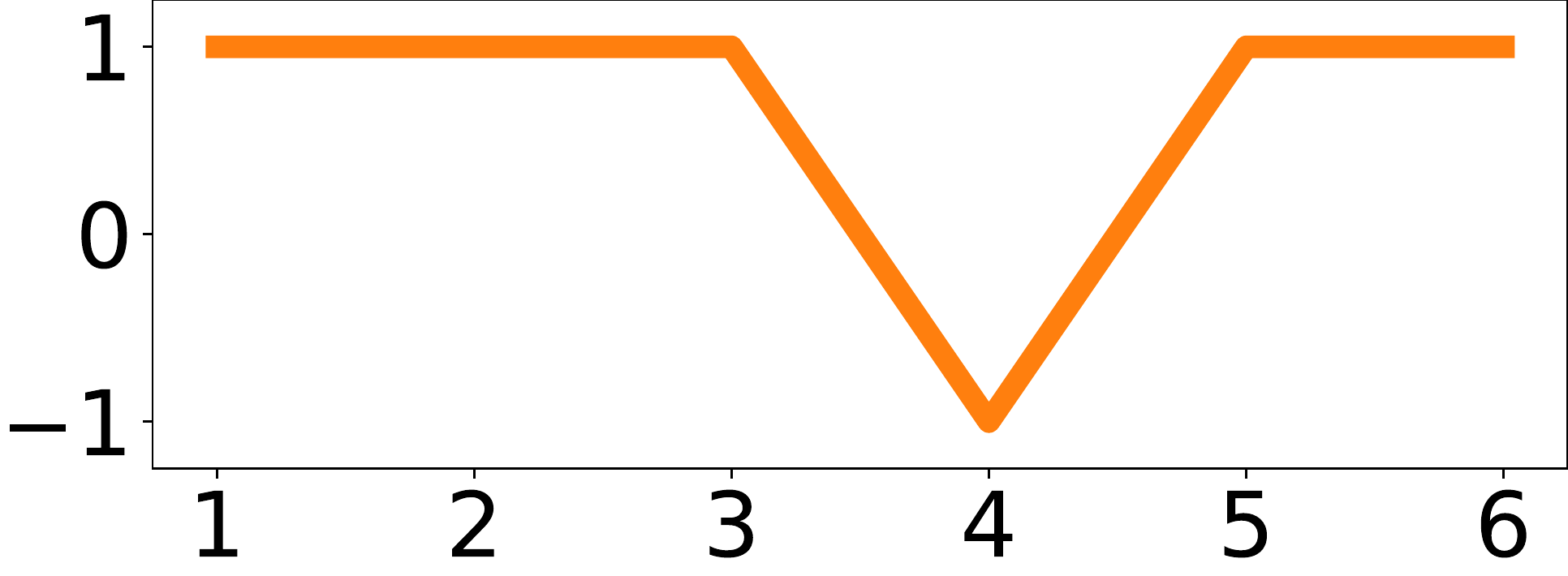}
        \caption{$T=\{1, 1, 1, -1, 1, 1\}$}
    \end{subfigure}
    \hfill
    \begin{subfigure}[b]{0.31\textwidth}
        \includegraphics[width=\textwidth]{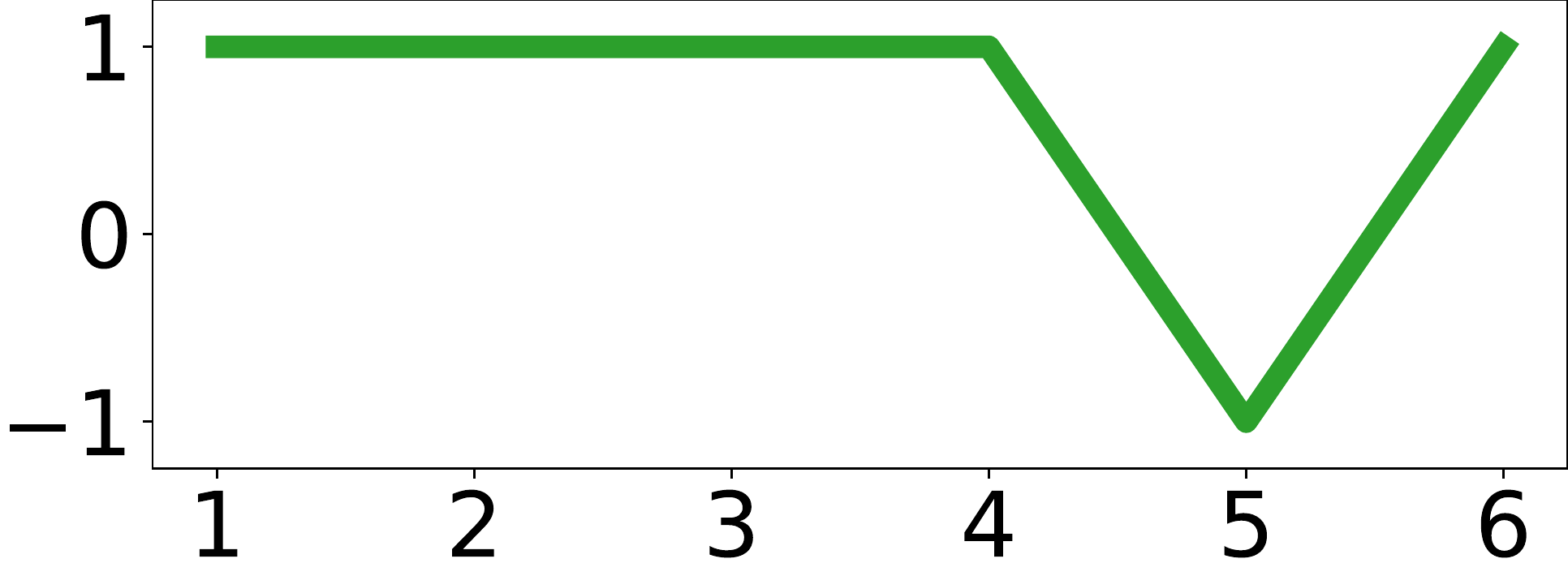}
        \caption{$U=\{1, 1, 1, 1, -1, 1\}$}
    \end{subfigure}
    \caption{\label{fig:3series}
    Three time series $S$, $T$ and $U$.
    No current variant of $\DTW$ captures the intuition that $S$ is identical to itself,
    and not identical to either $T$ or $U$,
    and that $S$ is more similar to $T$ than it is to $U$.}
\end{figure}

DTW sums pairwise-distances between aligned points in two series.
To allow for similar events that unfold at different rates,
$\DTW$ provides elasticity in which points are aligned.
However, it is well understood that $\DTW$ can be too flexible in the alignments it allows.
We illustrate this with respect to three series shown in Figure~\ref{fig:3series}.%
A natural expectation for any time series distance function $\similarity$
is that $\similarity(S,S) = 0 < \similarity(S,T) < \similarity(S,U)$.
However, $\DTW(S,S) = \DTW(S,T) = \DTW(S,U) = 0$.
While this may be appropriate for some tasks,
where all that matters is that these sequences all contain five 1s and one -1,
it is inappropriate for others.

Two variants have been developed to constrain $\DTW$'s flexibility,
Constrained $\DTW$ ($\CDTW$~\cite{sakoe1971, sakoe1978}, see Section~\ref{sec:relwork:cdtw}) and
Weighted $\DTW$ ($\WDTW$~\cite{jeongWeightedDynamicTime2011a}, see Section~\ref{sec:relwork:wdtw}).
$\CDTW_w$ constrains the alignments from warping by more than a window, $w$.
Within the window, any warping is allowed, and none beyond it.
Depending on $w$, there are three possibilities for how $\CDTW$ might assess our example:
\begin{enumerate}
    \item For $w\geq2$, $\CDTW_w(S,S) = \CDTW_w(S,T) = \CDTW_w(S,U) = 0$.
    \item $\CDTW_1(S,S) = \CDTW_1(S,T) = 0 < \CDTW_1(S,U)=8$ (see Figure~\ref{fig:3cdtw})
    \item $\CDTW_0(S,S) = 0 < \CDTW_0(S,T) = \CDTW_0(S,U)$. 
\end{enumerate}

$\WDTW$ applies a multiplicative penalty to alignments that increases the more the alignment is warped.
However, because the penalty is multiplicative, the perfectly matched elements in our example
still lead to $\WDTW(S,S) = \WDTW(S,T) = \WDTW(S,U) = 0$ (see Figure~\ref{fig:3wdtw}).

None of $\DTW$, $\WDTW$ or $\CDTW$ can be parameterized to obtain the natural expectation that
$\similarity(S,S) = 0 < \similarity(S,T) < \similarity(S,U)$.

In this paper, we present Amerced DTW ($\ADTW$), an intuitive, elegant and 
effective variant of $\DTW$ that applies a tunable additive penalty $\omega$ for warping an alignment.
To align the $-1$ in $S$ with the $-1$ in $T$,
it is necessary to warp the alignments by 1 step, incurring a cost of $\omega$.
A further compensatory warping step is required to bring the two end points back into alignment,
incurring another cost of $\omega$.
Thus, the total cost of aligning the two series is $\ADTW_\omega(S,T)=2\omega$.
Aligning the $-1$ in $S$ with the $-1$ in $U$ requires warping by two steps, incurring a penalty of $2\omega$,
with a further $2\omega$ penalty required to realign the end points, resulting in $\ADTW_\omega(S,U)=4\omega$.
Thus, $\ADTW$ can be parameterized to be in line with natural expectations for our example,
or to allow the flexibility to treat the three series as equivalent
if that is appropriate for an application:
\sloppy
\begin{enumerate}
    \item
    For $0<\omega<4$ (see Figure~\ref{fig:3adtw}),
    $\ADTW_\omega(S,S) = 0 < \ADTW_\omega(S,T) = 2\omega < \ADTW_\omega(S,U) = \min(4\omega,8)$.
    We have $\ADTW_\omega(S,U) = \min(4\omega,8)$
    because the path will not be warped if the resulting cost of $8$ is cheaper
    than 4 warping penalties.
    \item For $\omega\geq4$,
    ${\ADTW_\omega(S,S) = 0 < \ADTW_\omega(S,T) = \ADTW_\omega(S,U)=8}$,
    because the penalty for warping is greater than the cost of not warping.
    \item
    For $\omega=0$, ${\ADTW_0(S,S)= \ADTW_0(S,T) = \ADTW_0(S,U) = 0}$,
    because there is no penalty for warping.
\end{enumerate}
\fussy
We show that this approach is not only intuitive, it often provides superior outcomes in practice.

The remainder of this paper is organised as follows.
In Section~\ref{sec:relwork}, we review the literature related to DTW and its variants.
ADTW is presented Section~\ref{sec:adtw}, and Section~\ref{sec:params} discusses how to parameterize it.
We then present the results of our experiment in Section~\ref{sec:exp},
and conclude in Section~\ref{sec:conclusion}.

\section{\label{sec:relwork}Background and related Work}

$\DTW$ is a foundational technique for a wide range of time series data analysis tasks, including 
similarity search \cite{rakthanmanon2012searching},
regression \cite{tan2021regression},
clustering \cite{petitjean2011global},
anomaly and outlier detection \cite{diab2019anomaly},
motif discovery \cite{alaee2021time}, 
forecasting \cite{bandara2021improving},
and subspace projection \cite{DENG2020107210}.

In this paper, for ease of exposition, we only consider univariate time series,
although $\ADTW$ extends directly to the multivariate case.
We denote series by the capital letters $S$, $T$ and $U$.
The letter $\ell$ denotes the length of the series.
Subscripting (e.g. $\ell{}_S$) is used to disambiguate between the length of different series.
The elements $S_1, S_2, \dots S_i \dots{}S_{\ell{}_S}$ with $1\leq{}i\leq{}\ell{}_S$
are the elements of the series $S=(S_1, S_2, \dots S_i \dots S_{\ell{}_S})$,
drawn from a domain $\domain$ (e.g. real numbers).

We assess $\ADTW$ on time series classification tasks.
Given a training set $\Dtrain=\langle\Dtrain_1, \Dtrain_2, \ldots, \Dtrain_N\rangle$
of labeled time series $\Dtrain_i$,
time series classification learns a classifier which is then assessed
with respect to classifying an evaluation set $\Dtest=\langle\Dtest_1, \Dtest_2, \ldots, \Dtest_M\rangle$
of labeled time series.

\subsection{\label{sec:relwork:dtw}Dynamic Time Warping and Variants}
Dynamic Time Warping ($\DTW$) \cite{sakoe1971, sakoe1978}
handles alignment of series with distortion and disparate lengths.
An \emph{alignment}
$\Ali(S,T)=((i_1, j_1), \ldots, (i_{\alen}, j_{\alen}))$
between two series $S$ and $T$
is made of tuples $\Ali(S,T)_{1\leq{}k\leq{}\lambda}=(i_k, j_k)$
representing the point-to-point alignments of $S_{i_k}$ with $T_{j_k}$.
A \emph{warping path} is an alignment $\Ali(S,T)$ with the following properties:
\begin{itemize}
    \item Series extremities are aligned with each other, i.e.
        \[\Ali(S,T)_1=(1,1)\quad\text{and}\quad\Ali(S,T)_{\alen}=(\ell_S, \ell_T)\]
    \item It is continuous, i.e.
        \[\forall{k\in\{2,\alen\}}\ 
            (i_{k-1} \leq i_{k} \leq i_{k-1}+1)
            \ \wedge\ 
            (j_{k-1} \leq j_{k} \leq j_{k-1}+1)
        \]
    \item It is monotonic, i.e.
        \[\forall{k\in\{2,\alen\}}\  \Ali_k\neq\Ali_{k-1} \]
\end{itemize}
Given a cost function $\cost: \domain\times\domain\rightarrow\R$,
$\DTW$ finds a warping path $\Ali(S,T)$ minimizing the cumulative sum of $\cost(S_{i_k}, T_{j_k})$:
\begin{equation}\label{eq:dtw-align}
    \DTW(S,T)=min_{\Ali(S,T)}\sum_{k=1}^{\alen}\cost(S_{i_k},T_{j_k})
\end{equation}
In this paper, following common practice in time series classification,
we use the squared L2 norm as the cost function for all distances.

$\DTW(S,T)$ can be computed on a 0-indexed ${(\ell_S+1)\times(\ell_T+1)}$ \emph{cost matrix} $M_{\DTW(S,T)}$.
A cell $M_{\DTW(S,T)}(i,j)$ represents the minimal cumulative cost of aligning the first $i$ points of $S$
with the first $j$ points of $T$.
It follows that
\begin{equation*}
    \DTW(S,T)=M_{\DTW(S,T)}(\ell_S, \ell_T)
\end{equation*}

\begin{figure}[t]
    \centering
    \begin{subfigure}[b]{0.31\textwidth}
        \includegraphics[width=\textwidth]{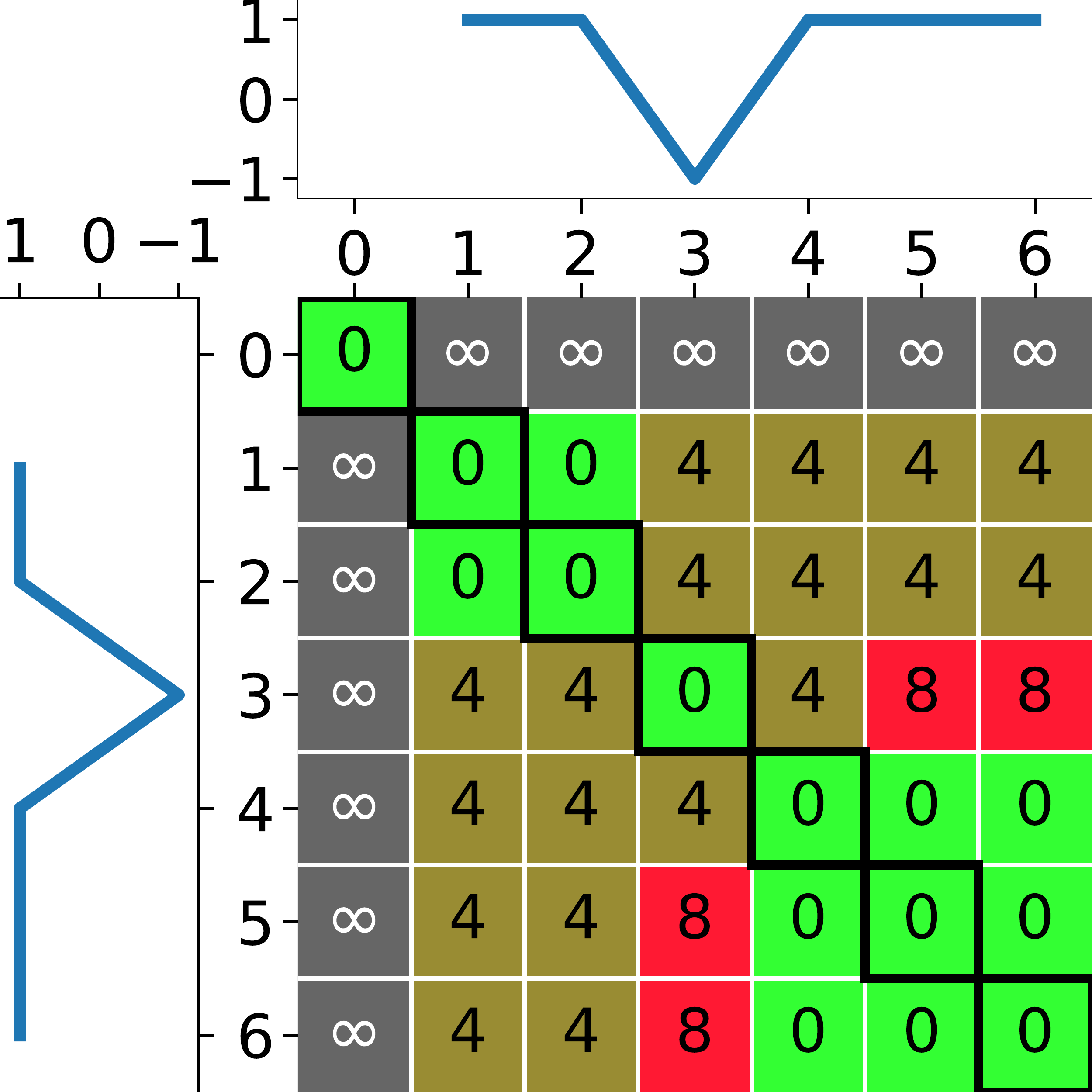}
        \caption{$\DTW(S,S)=0$}
    \end{subfigure}
    \hfill
    \begin{subfigure}[b]{0.31\textwidth}
        \includegraphics[width=\textwidth]{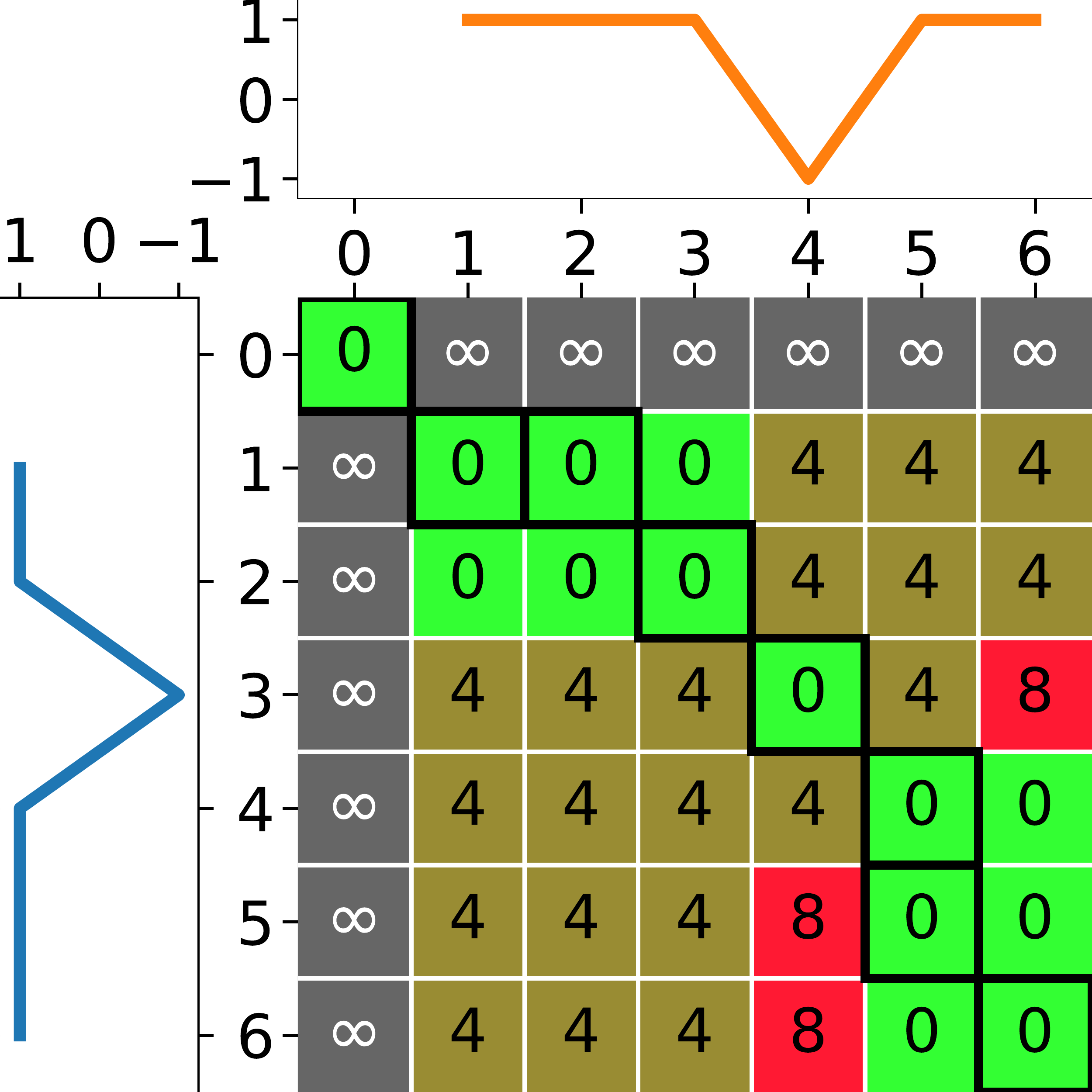}
        \caption{$\DTW(S,T)=0$}
    \end{subfigure}
    \hfill
    \begin{subfigure}[b]{0.31\textwidth}
        \includegraphics[width=\textwidth]{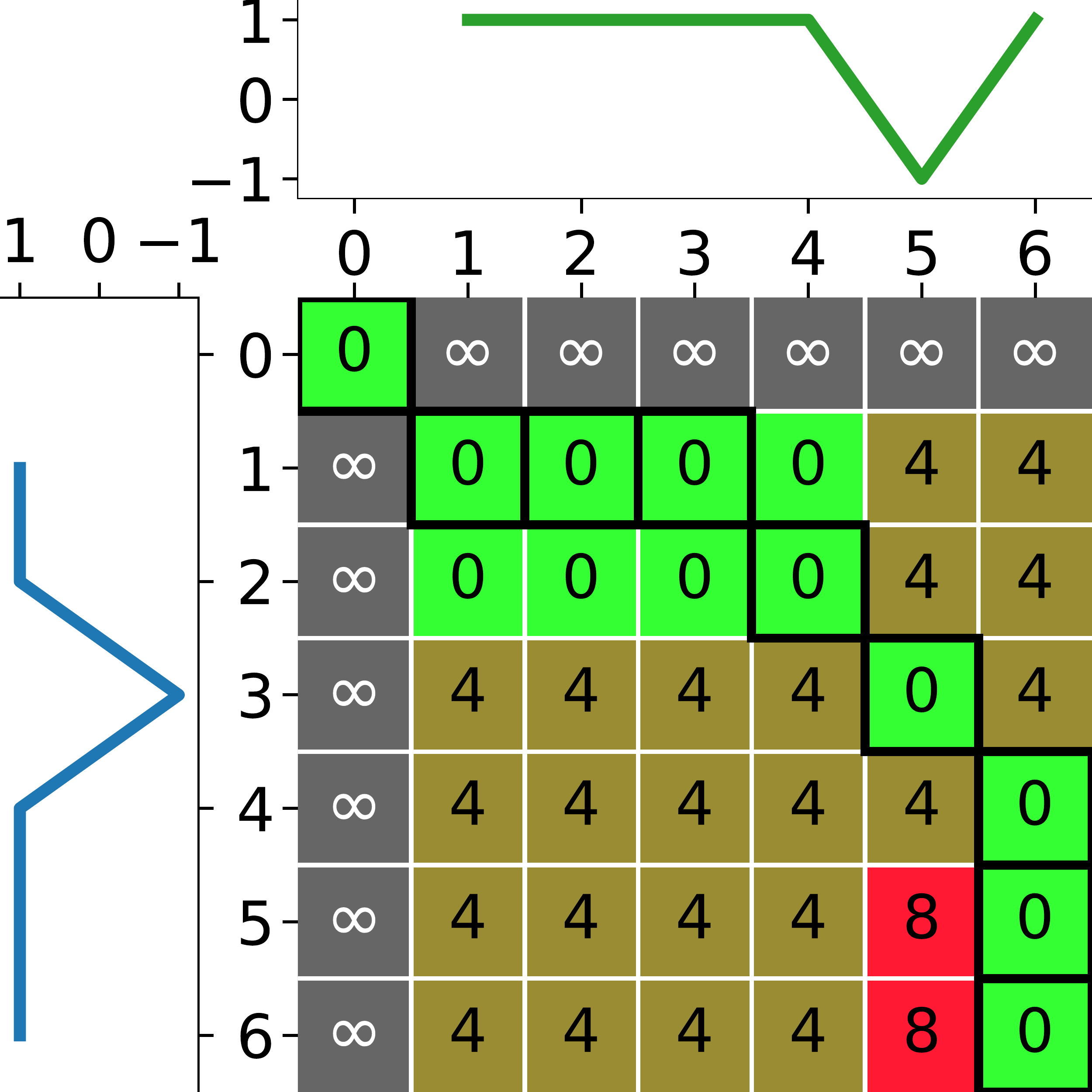}
        \caption{$\DTW(S,U)=0$}
    \end{subfigure}
    \caption{\label{fig:3dtw}
    Cost matrices and warping paths for
    $\DTW(S,S)$, $\DTW(S,T)$, and $\DTW(S,U)$.
    }
\end{figure}

The cost matrix $M_{\DTW(S,T)}$ is defined by a set of recursive equations~\eqref{eq:DTW}.
The first two equations~\eqref{eq:DTW:corner} and~\eqref{eq:DTW:border} defines the border conditions.
The third one~\eqref{eq:DTW:main} computes the cost of a cell $(i,j)$
by adding the cost of aligning $S_i$ with $T_j$,
given by $\cost(S_i, T_j)$, to the cost of its smallest predecessors.
Figure~\ref{fig:3dtw} shows examples of $\DTW$ cost matrices.
\begin{subequations}\label{eq:DTW}
    \begin{align}
        M_{\DTW(S, T)}(0,0) &= 0       \label{eq:DTW:corner}\\
        M_{\DTW(S, T)}(i,0) &= M_{\DTW(S, T)}(0,j) = +\infty \label{eq:DTW:border}\\
        M_{\DTW(S, T)}(i,j) &= \cost(S_i, T_j) + \min\left\{
        \begin{aligned}
            &M_{\DTW(S, T)}(i-1, j-1) \\
            &M_{\DTW(S, T)}(i-1, j) \\
            &M_{\DTW(S, T)}(i, j-1)
        \end{aligned}
        \right. \label{eq:DTW:main}
    \end{align}
\end{subequations}

An efficient implementation technique~\cite{herrmann2021early}
allows $\DTW$ and its variants, including $\ADTW$,
to be computed with an $O(\ell)$ space complexity, and a \emph{pruned} time complexity;
the worst case scenario $O(\ell^2)$ can usually be avoided.

\subsection{\label{sec:relwork:cdtw}Constrained DTW}
\begin{figure}[t]
    \centering
    \begin{subfigure}[b]{0.31\textwidth}
        \includegraphics[width=\textwidth]{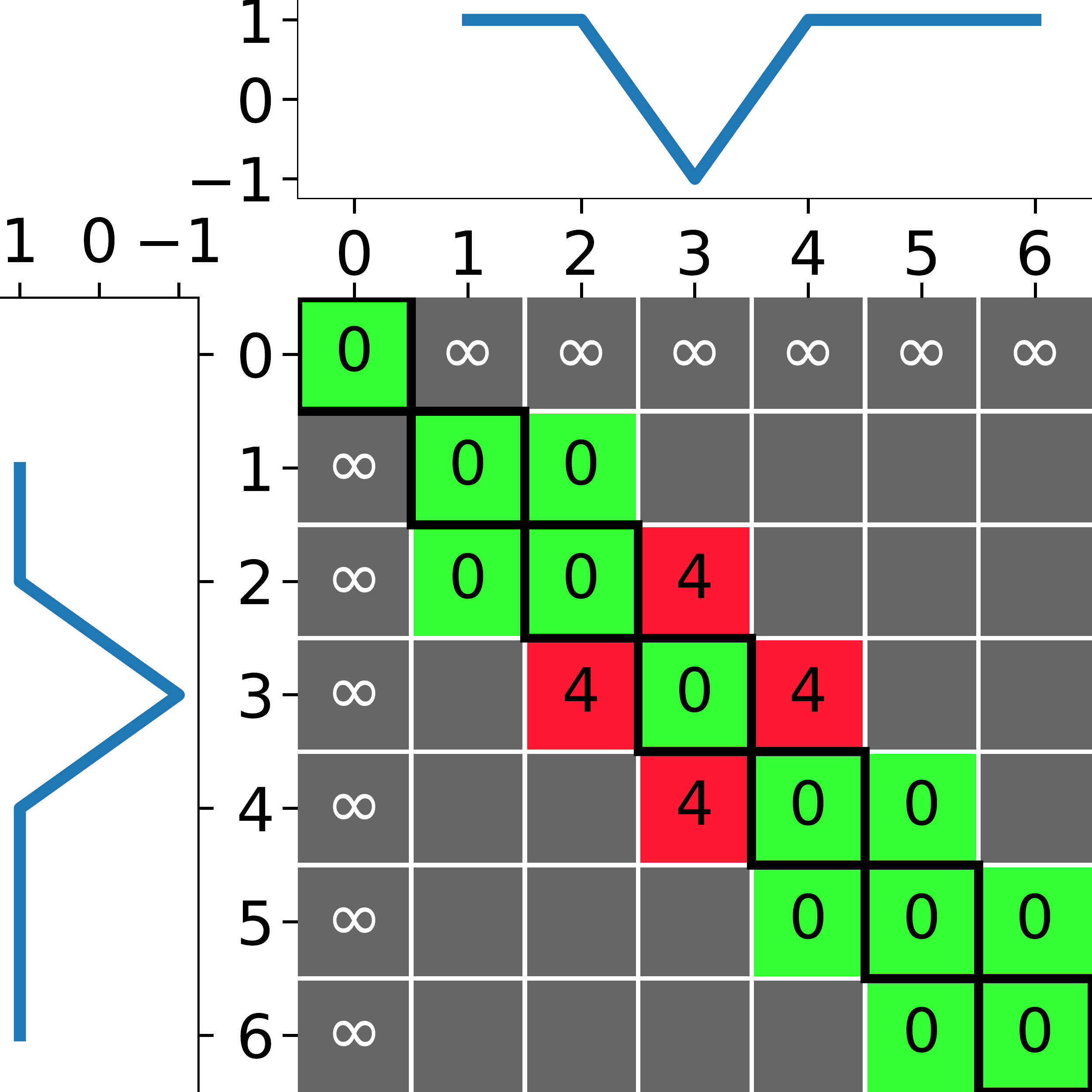}
        \caption{$\CDTW_1(S,S)=0$}
    \end{subfigure}
    \hfill
    \begin{subfigure}[b]{0.31\textwidth}
        \includegraphics[width=\textwidth]{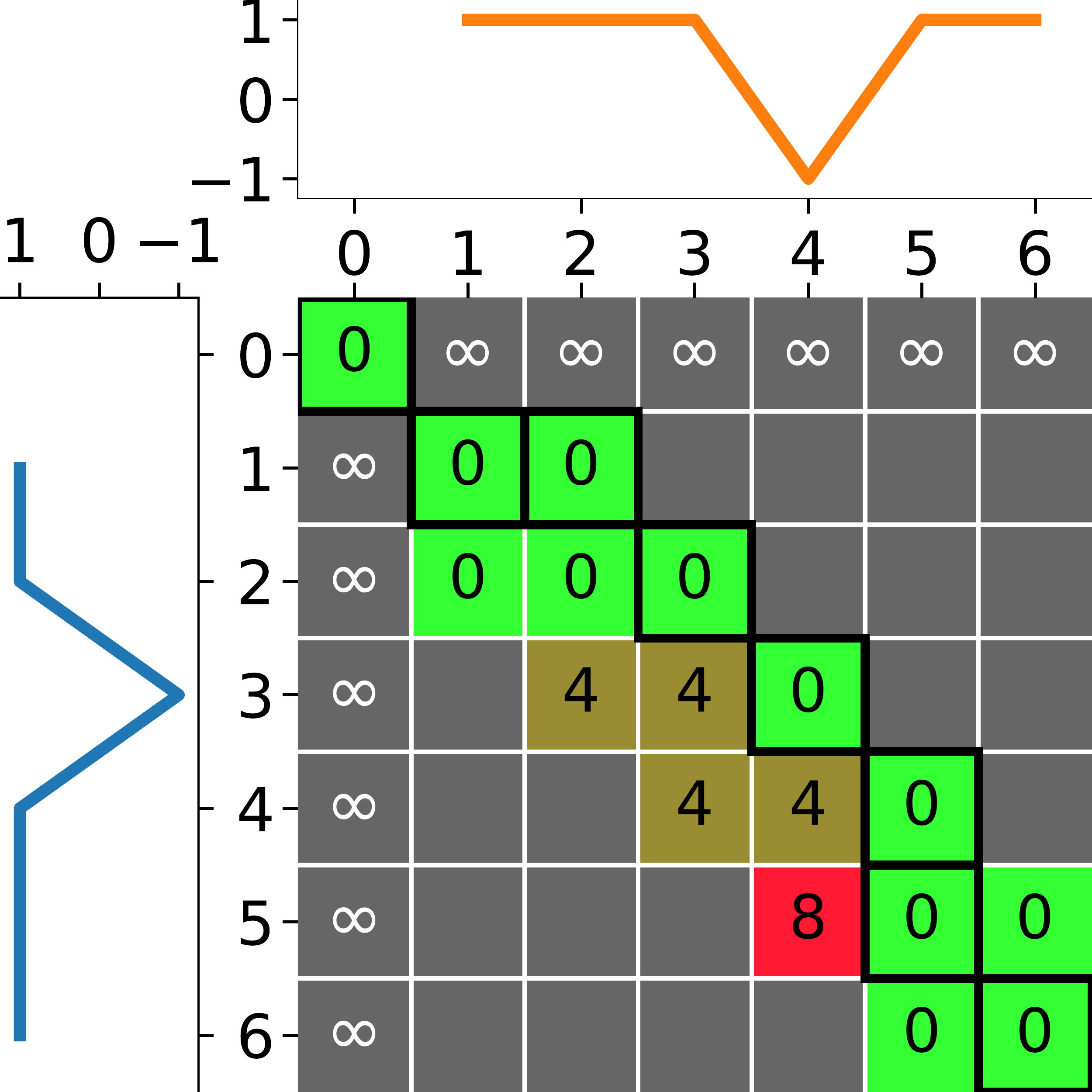}
        \caption{$\CDTW_1(S,T)=0$}
    \end{subfigure}
    \hfill
    \begin{subfigure}[b]{0.31\textwidth}
        \includegraphics[width=\textwidth]{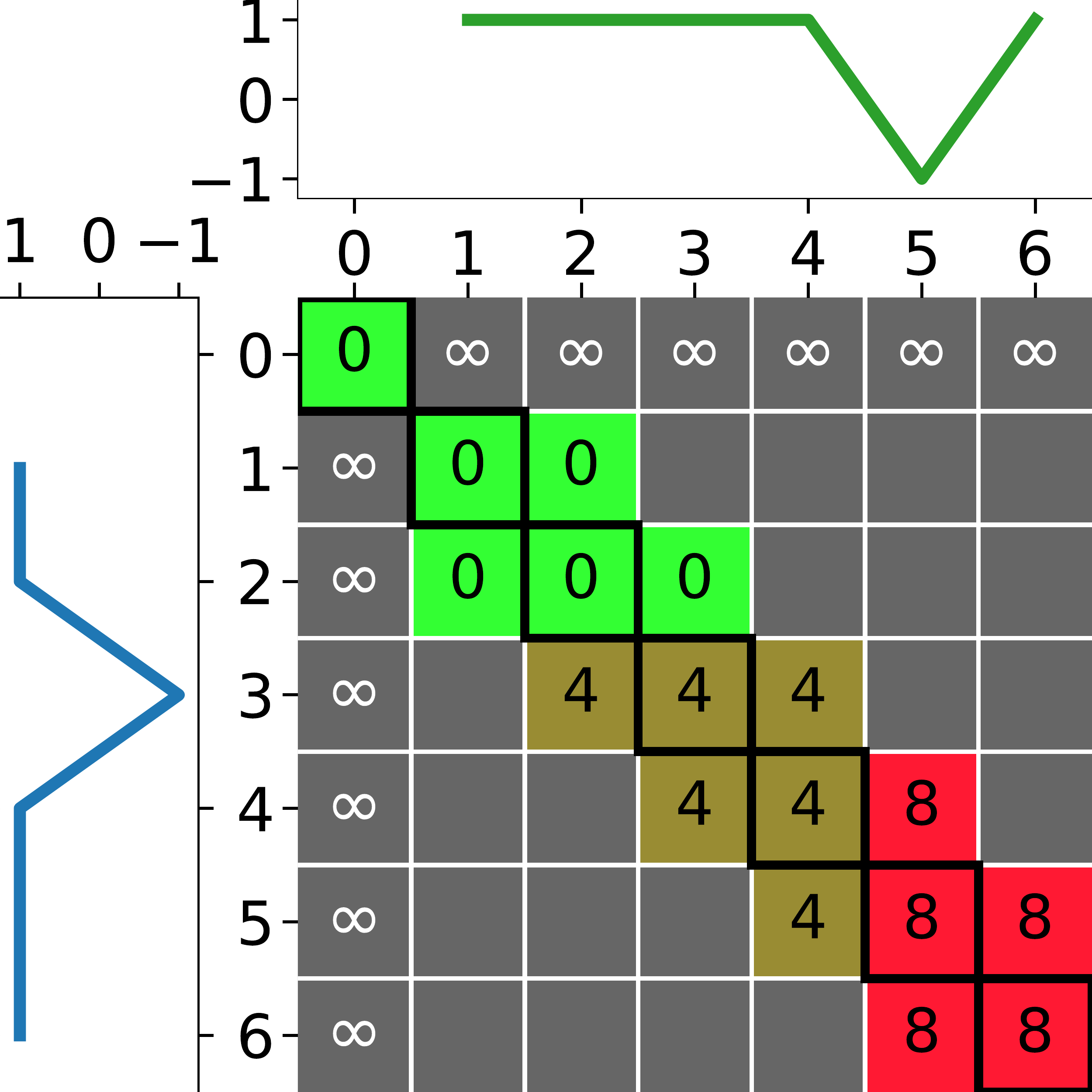}
        \caption{$\CDTW_1(S,U)=8$}
    \end{subfigure}
    \caption{\label{fig:3cdtw}
    Cost matrices and warping paths for
    $\CDTW_1(S,S)$, $\CDTW_1(S,T)$, and $\CDTW_1(S,U)$.}
\end{figure}

In Constrained DTW ($\CDTW$), the warping path is strictly restricted to a subarea of the cost matrix.
Different constraints, with different subarea ``shapes'', 
exist~\cite{sakoe1978,itakuraMinimumPredictionResidual1975}.
We focus on the popular Sakoe-Chiba band~\cite{sakoe1978},
also known as a \emph{warping window} (or \emph{window} for short).
The window is a parameter $w$ controlling how far the warping path can deviate from the diagonal.
Given a cost matrix $M_{\CDTW_w(S,T)}$,
a line index $1\leq{}i\leq{}\ell{}$ and a column index $1\leq{}j\leq{}\ell{}$,
we constrain $(i,j)$ by $\abs{i-j}\leq{}w$.
A window of $0$ forces the warping path to be on the diagonal,
while a window of $\ell{}-1$ is equivalent to $\DTW$ (no constraint). 
For example with $w=1$,
the warping path can only step one cell away from each side of the diagonal (Figure~\ref{fig:3cdtw}).

An effective window increases the usefulness of $\CDTW$ over $\DTW$ by preventing spurious alignments.
For example, nearest neighbor classification using $\CDTW$ as the distance measure (NN-$\CDTW$)
is significantly more accurate than nearest neighbor classification using $\DTW$
(NN-$\DTW$, see Section~\ref{sec:exp}).
$\CDTW$ is also faster to compute than $\DTW$, as cells of the cost matrix beyond the window are ignored.
However, the window is a parameter that must be learned.
The Sakoe-Chiba band is part of the original definition of $\DTW$ and the defacto default constraint for $\CDTW$
(e.g. in~\cite{lines2015,lucasProximityForestEffective2019,shifazTSCHIEFScalableAccurate2020}).

One of the issues that must be addressed in any application of $\CDTW$
is how to select an appropriate value for $w$.
A method deployed when the UCR benchmark archive was first established has become the defacto
default method for time series classification \cite{dauUCRTimeSeries2019}.
This method applies leave-one-out cross validation to nearest neighbor classification
over the training data $\Dtrain$ for all $w\in\{0, 0.01\ell, 0.02\ell,\ldots,\ell\}$.
The best value of $w$ is selected, with respect to the performance measure of interest
(typically the lowest error).

\subsection{\label{sec:relwork:wdtw}Weighted DTW}
\begin{figure}
    \centering
    \begin{subfigure}[b]{0.31\textwidth}
        \includegraphics[width=\textwidth]{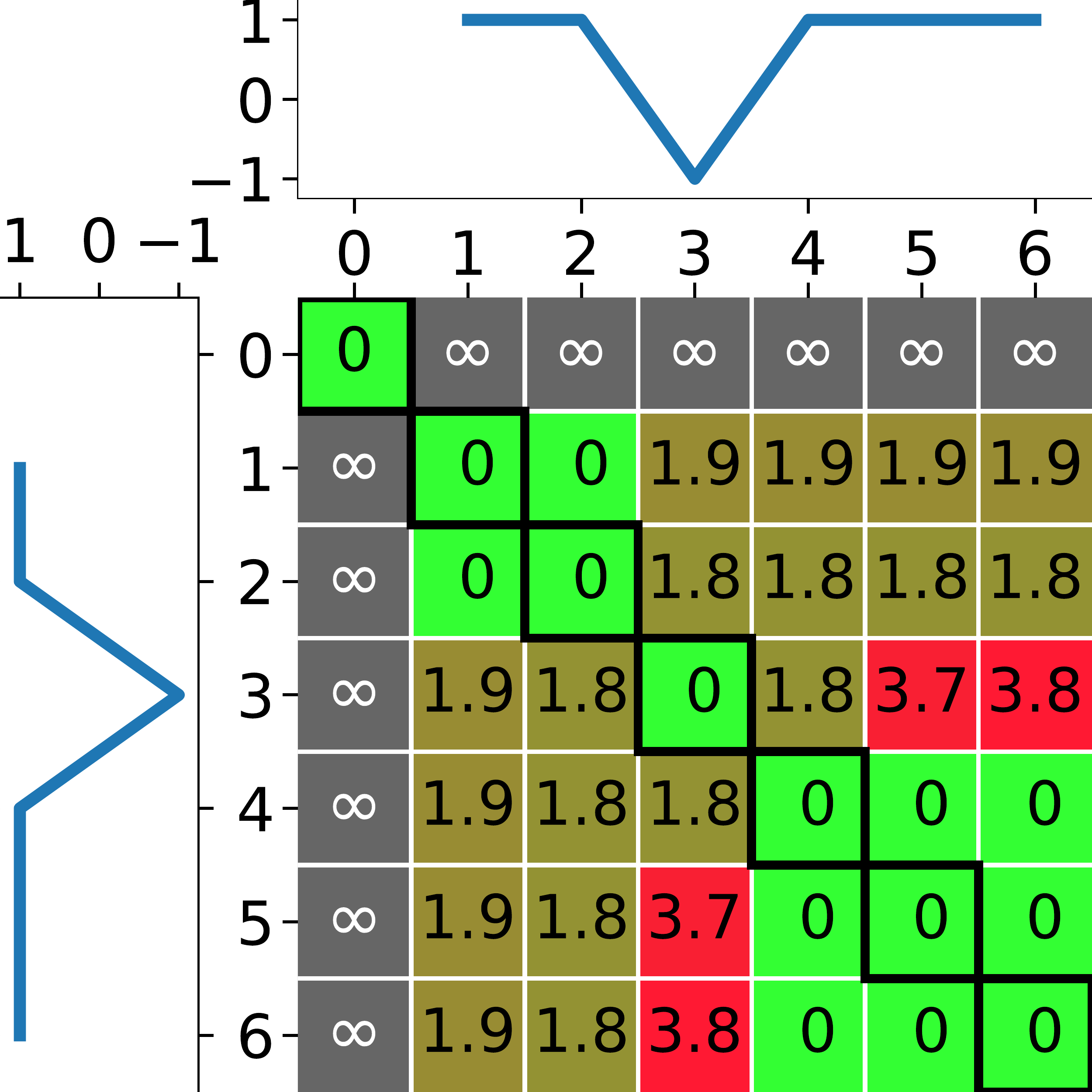}
        \caption{$\WDTW_{0.1}(S,S)$}
    \end{subfigure}
    \hfill
    \begin{subfigure}[b]{0.31\textwidth}
        \includegraphics[width=\textwidth]{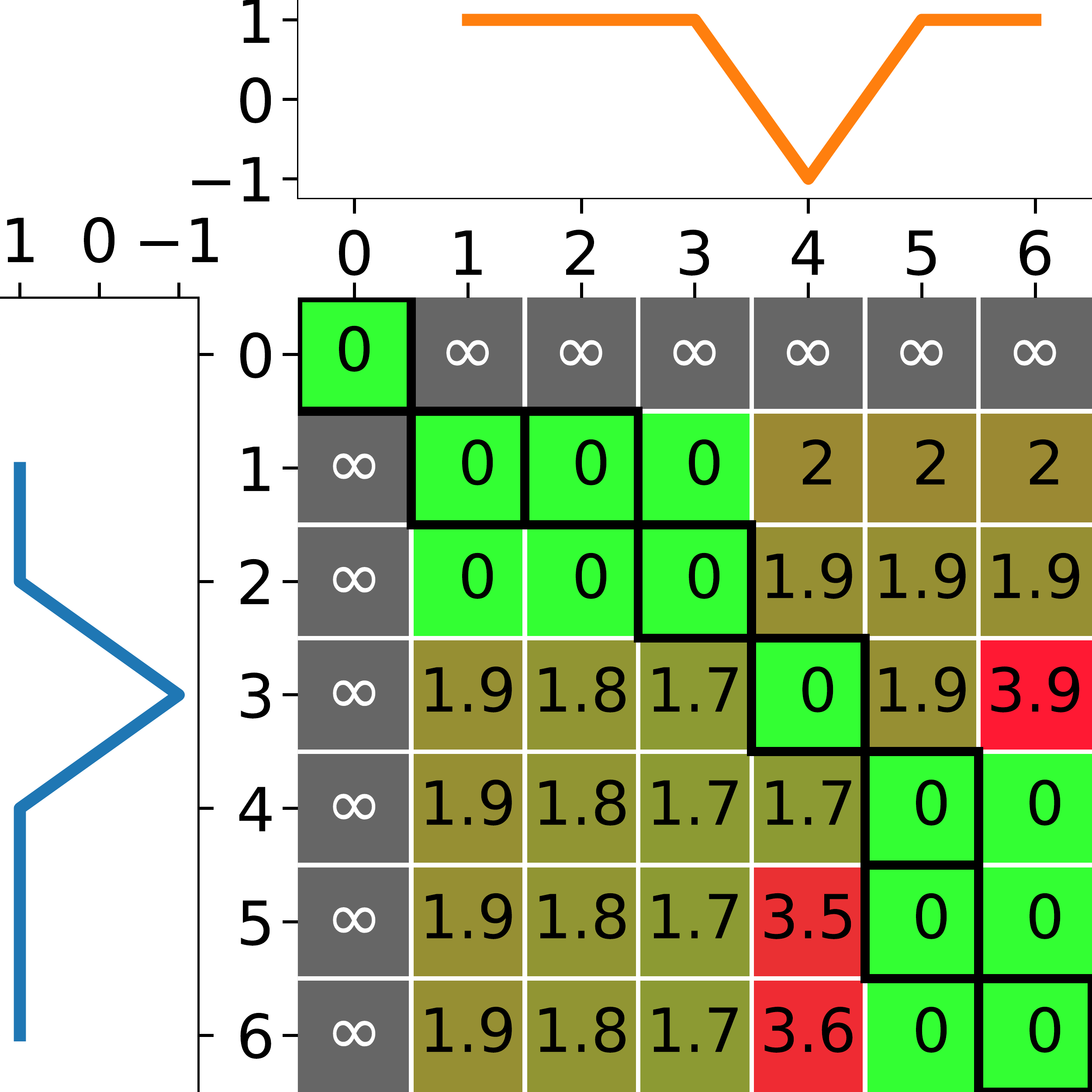}
        \caption{$\WDTW_{0.1}(S,T)$}
    \end{subfigure}
    \hfill
    \begin{subfigure}[b]{0.31\textwidth}
        \includegraphics[width=\textwidth]{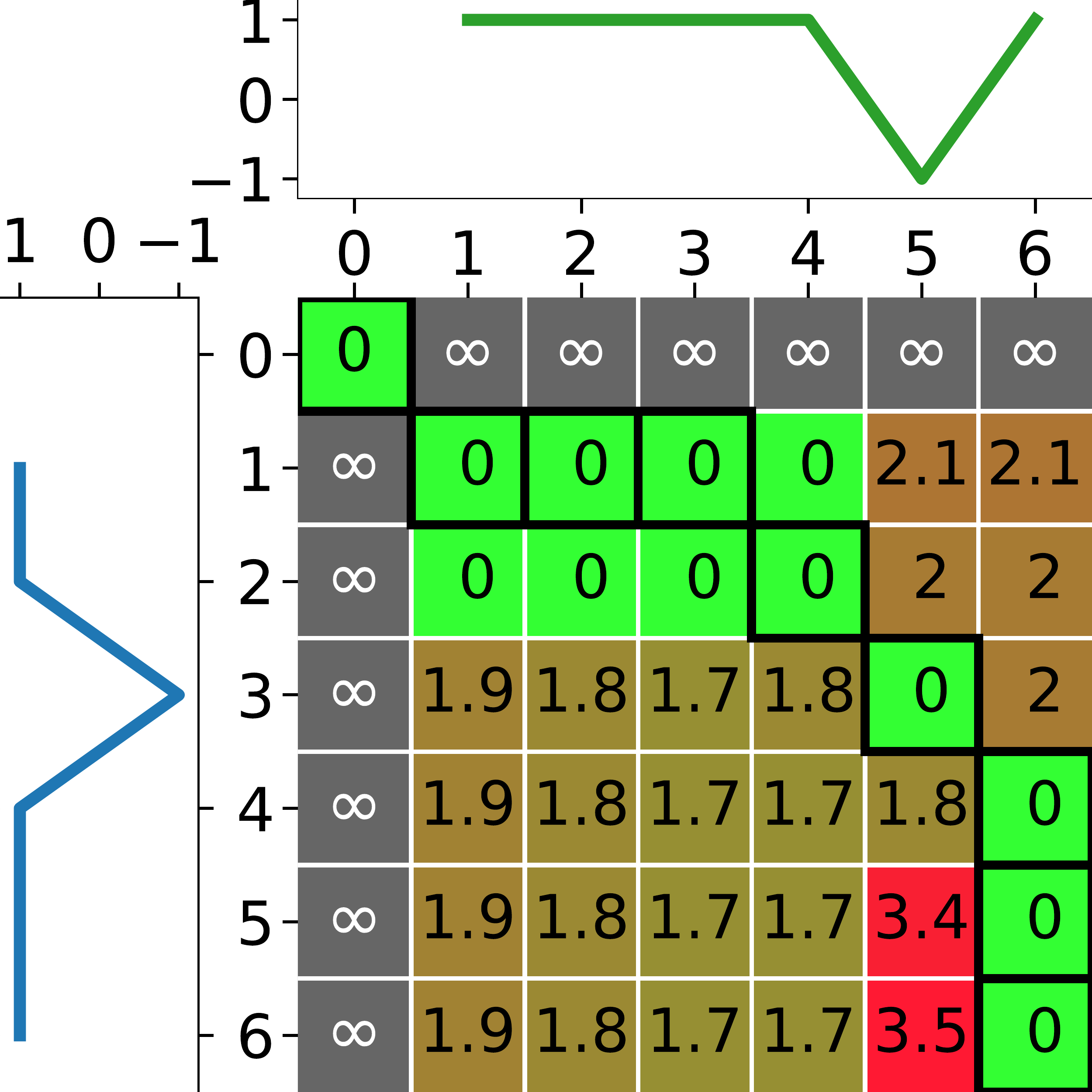}
        \caption{$\WDTW_{0.1}(S,U)$}
    \end{subfigure}
    \caption{\label{fig:3wdtw}
        Cost matrices and warping paths 
        for $\WDTW_{0.1}(S,S)$, $\WDTW_{0.1}(S,T)$, and $\WDTW_{0.1}(S,U)$.
    }
\end{figure}

Weighted Dynamic Time Warping ($\WDTW$)~\cite{jeongWeightedDynamicTime2011a}
penalizes phase difference between two points.
An alignment cost of a cell $(i, j)$ is weighted according to its distance to the diagonal, $\delta=\abs{i-j}$.
In other words, $\WDTW$ relies on a $\weight$~\eqref{eq:WDTW:w} function
to define a new cost function $\cost'$~\eqref{eq:WDTW:cost}.
A large weight decreases the chances of a cell to be on an optimal path.
The \emph{weight factor} parameter $g$ controls the penalization,
and usually lies within $0.01$ -- $0.6$~\citep{jeongWeightedDynamicTime2011a}.
Figure~\ref{fig:3wdtw} shows several $\WDTW$ cost matrices.
\begin{subequations}
    \begin{align}
    \weight(\delta)  &= \frac{1}{1+\exp^{-g\times{}(\delta-\ell{}/2)}} \label{eq:WDTW:w}\\
    \cost'(S_i, T_j) &= \cost(S_i, T_j)*\weight(\abs{i-j}) \label{eq:WDTW:cost}
    \end{align}
\end{subequations}

$\WDTW$ applies a penalty to every pair of aligned points that are off the diagonal.
Hence, the longer the off diagonal path is, the greater the penalty.
Thus, it favors many small deviations from the diagonal over fewer longer deviations.

$\WDTW$ is also confronted with the problem of selecting an appropriate value for its parameter, $g$.
We use the method employed in the Elastic Ensemble \cite{lines2015},
and modeled on the defacto method for parameterizing $\CDTW$.
This method applies leave-one-out cross validation to nearest neighbor classification
over the training data $\Dtrain$ for all $g\in\{0.01, 0.02, \ldots1.0\}$.

\subsection{\label{sec:relwork:sqed}Squared Euclidean Distance}
One simple distance measure is to simply sum the cost of aligning successive points in the two series.
This is equivalent to $\CDTW$ with $w=0$,
and is only defined for same length series.
When the cost function for aligning two points $S_i$ and $T_j$ is $\cost(S_i,T_j)=(S_i-T_j)^2$,
this distance measure is known as the Squared Euclidean Distance:
\begin{equation}
    \SQED(S,T) = \sum_{i=1}^{\ell}(S_i-T_i)^2
\end{equation}

\section{\label{sec:adtw}Amerced Dynamic Time Warping}
Amerced Dynamic Time Warping ($\ADTW$) provides a new, intuitive, elegant,
and effective mechanism for constraining the alignments within the $\DTW$ framework.
It achieves this by the introduction of a novel constraint, \emph{amercing}
--- the application of a simple additive penalty $\omega$ every time an alignment is warped ($i-j$ changes).
This addresses the following problems:
\begin{itemize}
    \item $\DTW$ can be too flexible in its alignments;
    \item $\CDTW$ uses a crude step function ---
            any flexibility is allowed within the window and none beyond it;
    \item $\WDTW$ applies a multiplicative weight, and hence promotes large degrees of warping
                if they lead to low cost alignments ---
                the penalty incurred for warping is dependent on the costs of the warped alignments.
                Further, as the penalty is paid for every off-diagonal alignment (where $i\neq j$),
                $\WDTW$ penalizes off diagonal paths for their length,
                rather than for the number of times they adjust the alignment
                \footnote{By \emph{adjust the alignment} we mean having successive alignments $\Ali_k$ and $\Ali_{k+1}$ such that $i_{k+1}-i_k\neq j_{k+1}-j_k$.}.
\end{itemize}

\subsection{\label{sec:adtw:definition}Formal definition of ADTW}
Given two times series $S$ and $T$,
a cost function $\cost:\domain\times\domain\rightarrow\R$,
and an amercing penalty $\omega\in\R$ (see Section~\ref{sec:params}),
$\ADTW$ finds a warping path $\Ali(S,T)$ minimizing the cumulative sum of amerced costs:
\begin{multline}\label{eq:adtw-align}
    \ADTW_\omega(S,T)=\\
    \min_{\Ali(S,T)}\left[
    \cost(S_{i_1},T_{j_1})
    +\sum_{k=2}^{\alen}\cost(S_{i_k},T_{j_k})+1(i_{k}-i_{k-1}\neq j_{k}-j_{k-1})\omega
    \right]
\end{multline}
where $1(i_{k}-i_{k-1}\neq j_{k}-j_{k-1})\omega$ indicates that the amercing penalty $\omega$ 
is only applied if a step from $(i_{k-1},j_{k-1})$ to $(i_{k},j_{k})$ is not an increment on both series,
i.e. it is \emph{not} the case that $i_{k}=i_{k-1}+1$ and $j_{k}=j_{k-1}+1$.
$\ADTW_\omega(S.T)$ can be computed on a cost matrix $M_{\ADTW_\omega(S,T)}$~\eqref{eq:ADTW} with
\begin{equation*}
    \ADTW_\omega(S,T) = M_{\ADTW_\omega(S,T)})(\ell_S, \ell_T).
\end{equation*}

$M_{\ADTW}$ has the same border conditions as $M_{\DTW}$
(equations~\eqref{eq:ADTW:corner} and~\eqref{eq:ADTW:border}).
The other matrix cells are computed recursively,
\emph{amercing} the off diagonal alignments by the penalty $\omega$~\eqref{eq:ADTW:main}.
Methods for choosing $\omega$ are discussed Section~\ref{sec:params}.
\begin{subequations}\label{eq:ADTW}
    \begin{align}
        M_{\ADTW_\omega(S, T)}(0,0) &= 0       \label{eq:ADTW:corner}\\
        M_{\ADTW_\omega(S, T)}(i,0) &= M_{\ADTW_\omega(S, T)}(0,j) = +\infty \label{eq:ADTW:border}\\
        M_{\ADTW_\omega(S, T)}(i,j) &=  \min\left\{
        \begin{aligned}
            &M_{\ADTW_\omega(S, T)}(i-1, j-1) + \cost(S_i, T_j) \\
            &M_{\ADTW_\omega(S, T)}(i-1, j) + \cost(S_i, T_j) +\omega \\
            &M_{\ADTW_\omega(S, T)}(i, j-1) + \cost(S_i, T_j) +\omega
        \end{aligned}
        \right. \label{eq:ADTW:main}
    \end{align}
\end{subequations}

\begin{figure}
    \centering
    \begin{subfigure}[b]{0.31\textwidth}
        \includegraphics[width=\textwidth]{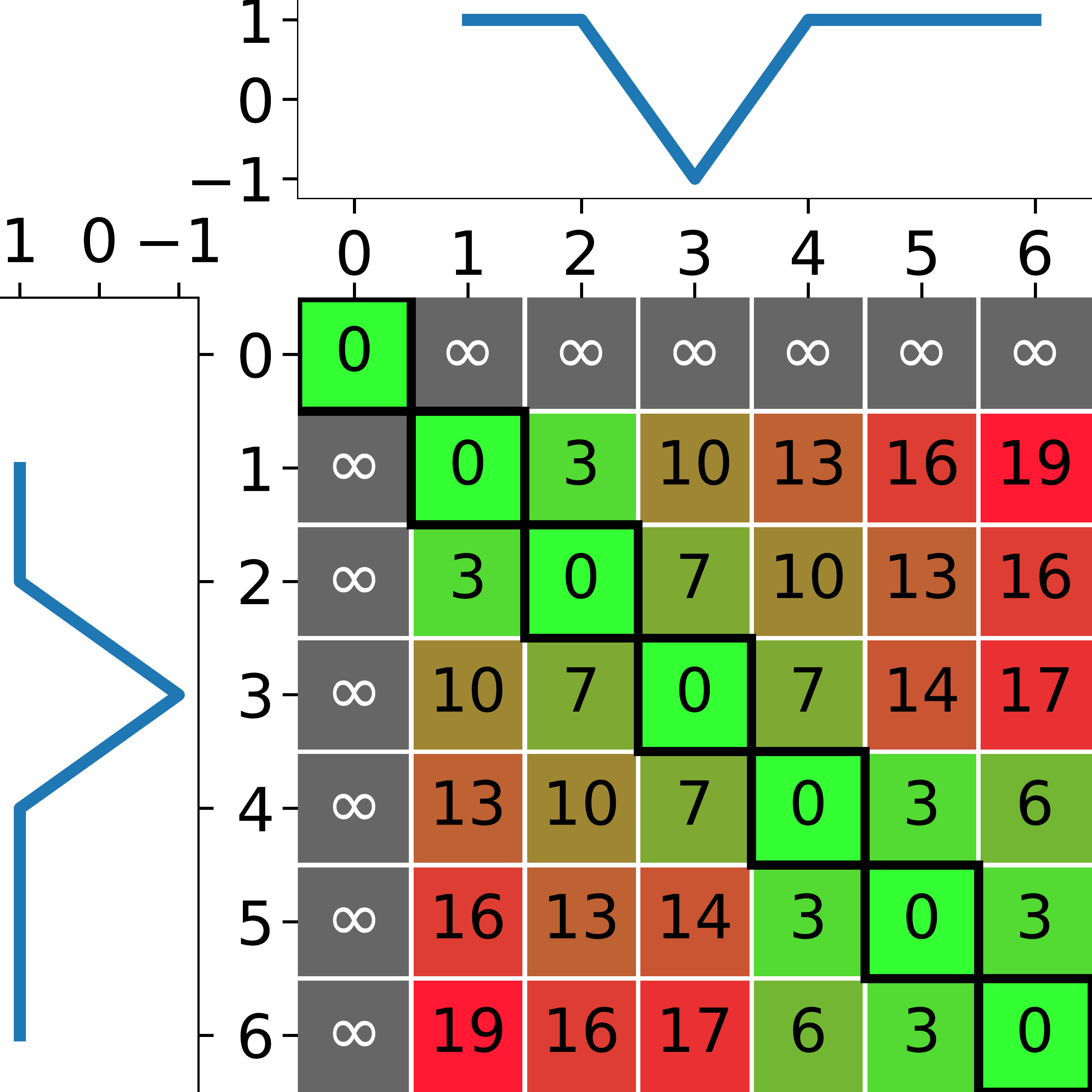}
        \caption{\label{fig:3adtw:a}$\ADTW_3(S,S)$}
    \end{subfigure}
    \hfill
    \begin{subfigure}[b]{0.31\textwidth}
        \includegraphics[width=\textwidth]{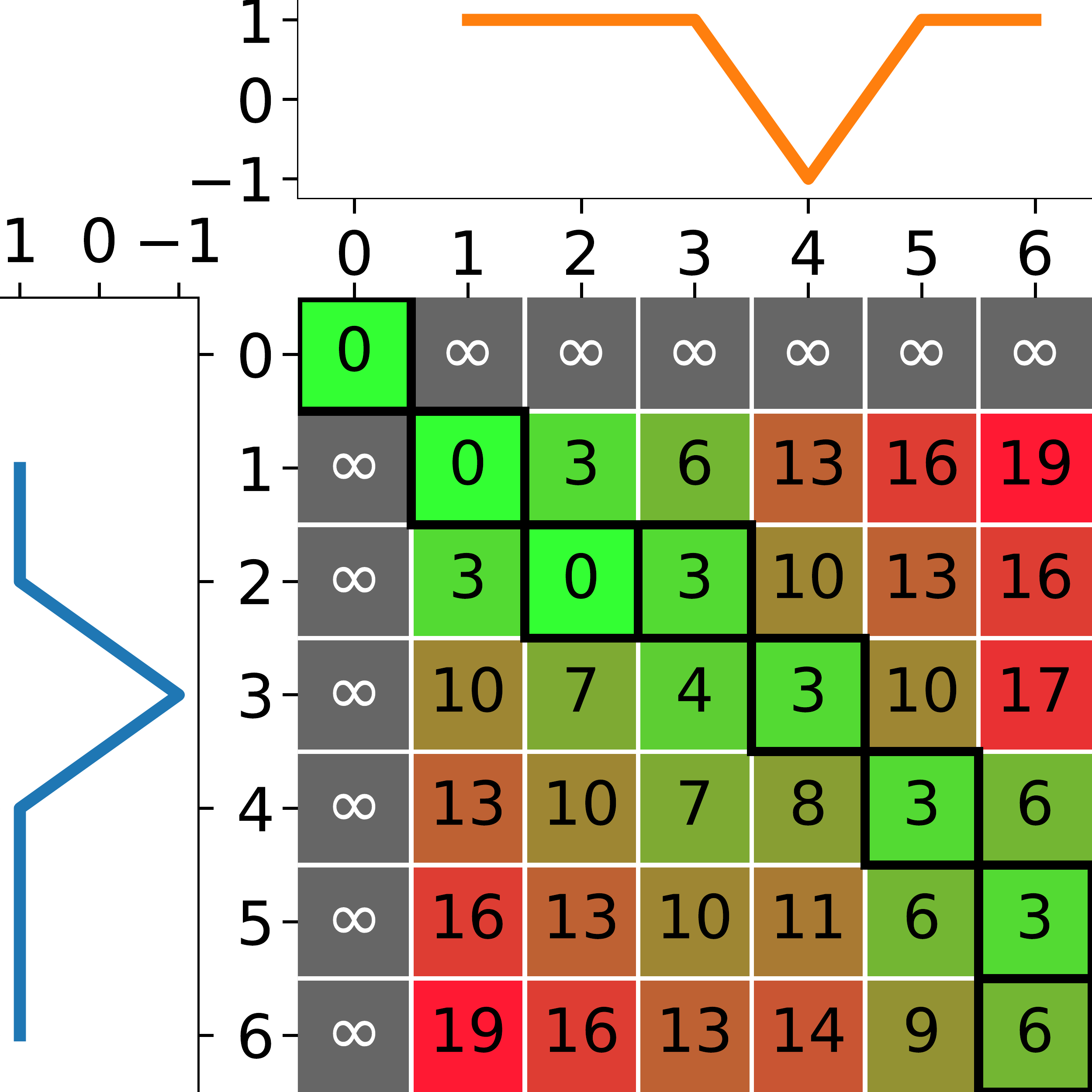}
        \caption{\label{fig:3adtw:b}$\ADTW_3(S,T)$}
    \end{subfigure}
    \hfill
    \begin{subfigure}[b]{0.31\textwidth}
        \includegraphics[width=\textwidth]{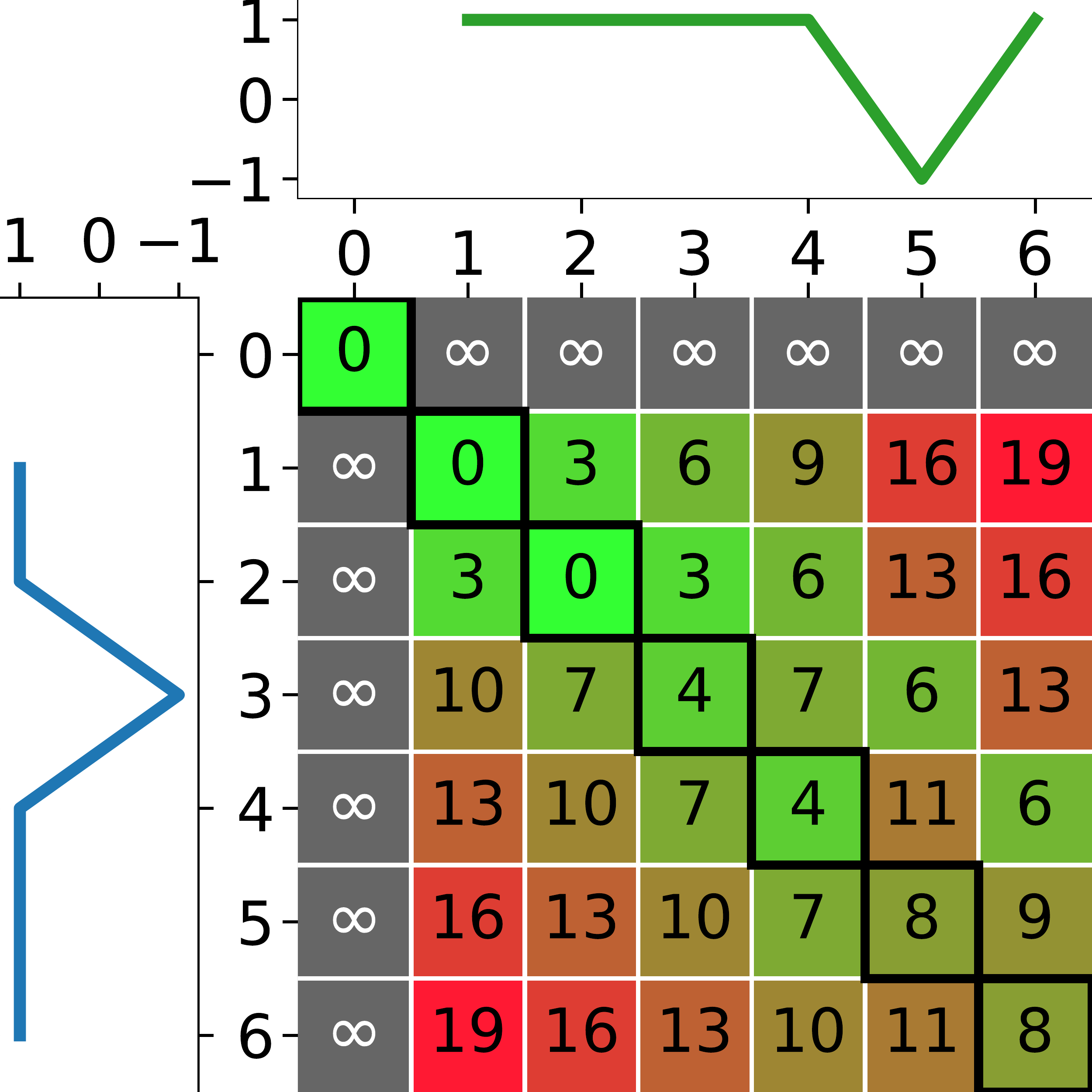}
        \caption{\label{fig:3adtw:c}$\ADTW_3(S,U)$}
    \end{subfigure}
    \caption{\label{fig:3adtw}
        Cost matrices and warping paths for $\ADTW_3(S,S)$, $\ADTW_3(S,T)$, and $\ADTW_3(S,U)$.
    }
\end{figure}

\subsection{\label{sec:adtw:properties}Properties of ADTW}
$\ADTW_{\omega}(S,T)$ is monotonic with respect to $\omega$.
\begin{equation}\label{eq:monotonic}
    \alpha<\beta \equiv \ADTW_{\alpha}(S,T) \leq \ADTW_{\beta}(S,T)
\end{equation}
\begin{proof}
    For any warping path $\Ali(S,T)$ that minimizes \eqref{eq:adtw-align} under $\omega=\beta$,
    the same path will result in an equivalent or lower score for $\omega=\alpha$.
    Hence the minimum score for a warping path under $\alpha$
    cannot exceed the minimum score under $\beta$.
\end{proof}

\begin{equation}\label{eq:dtw-bound}
    \ADTW_0(S,T) = \DTW(S,T)
\end{equation}
\begin{proof}
    With the amercing term $\omega$ set to zero,
    \eqref{eq:adtw-align}  is equivalent to \eqref{eq:dtw-align}.
\end{proof}

If $\ell_S=\ell_T$,
\begin{equation}\label{eq:sqed-bound}
    \ADTW_\infty(S,T) = \SQED(S,T)
\end{equation}
\begin{proof}
    With $\omega=\infty$,
    any off diagonal alignment will receive an infinite penalty.
    Hence the minimal cost warping path must follow the diagonal.
\end{proof}

When $0\leq\omega\leq\infty$, 
\begin{equation}\label{eq:adtw-bounds}
    \DTW(S,T)\leq\ADTW_\omega(S,T)\leq\SQED(S,T)
\end{equation}
\begin{proof}
    This follows from \eqref{eq:monotonic}, \eqref{eq:dtw-bound} and \eqref{eq:sqed-bound}.
\end{proof}
This observation provides useful and intuitive bounds for the measure.

$\ADTW$ is symmetric with respect to the order of the arguments. 
\begin{equation}\label{eq:adtw-sym}
    \ADTW_\omega(S,T) = \ADTW_\omega(T,S)
\end{equation}
\begin{proof}
    If we consider the cost matrix, such as illustrated in Figure~\ref{fig:3adtw},
    swapping $S$ for $T$ has the effect of flipping the matrix on the diagonal.
    This does not affect the cost of the minimal path.
\end{proof}
That a distance measure should satisfy this symmetry is intuitive,
as it does not seem appropriate that the order of the arguments
to a distance function should affect its value.
This symmetry also holds for the other variants of $\DTW$. 

$\ADTW$ is symmetric with respect to the order in which the series are processed.
With $\reverse(S)=(S_{\ell_S}, S_{\ell_S-1}, \ldots, S_1)$, we have:
\begin{equation}
    \ADTW_\omega(S,T) = \ADTW_\omega(\reverse(S), \reverse(T))
\end{equation}
\begin{proof}
For any warping path $\Ali(S,T)=((i_1, j_1), \ldots, (i_{\alen}, j_{\alen}))$,
there is a matching $\Ali(\reverse(S),\reverse(T))=((i_{\lambda},j_{\lambda}),\ldots,(i_1,j_1))$.
As these warping paths both contain the same alignments, the cost terms $\cost$ will be the same.
As $1(i_k{-}i_{k-1}\neq j_k{-}j_{k-1})\omega=1(i_{k-1}{-}i_{k}\neq j_{k-1}{-}j_{k})\omega$,
the amercing penalty terms will also be identical.
\end{proof}

This symmetry is also intuitive,
as it does not seem appropriate that the direction from which one calculates
a distance measure should affect its value.
Neither $\CDTW$ nor $\WDTW$ has this symmetry when $\ell_S\neq\ell_T$.
This is because both measures apply constraints relative to the diagonal,
which is different depending on whether one traces it from the top left or the bottom right of the cost matrix.
Unlike $\CDTW(S,T)$, $\ADTW(S,T)$ is well defined when $\ell_S\neq\ell_T$.
If $w<|\ell_S-\ell_T|$ then $\CDTW_w(S,T)$ is undefined.

\section{\label{sec:params}Parameterization}
As shown Section~\ref{sec:adtw:properties},
$\ADTW_{\omega}(S,T)$ can be parameterized so as to range from being as flexible as $\DTW$,
to being as constrained as $\SQED$.
If the situation requires a large amount of warping, a \emph{small} penalty should be used.
Reciprocally, a \emph{large} penalty helps to avoid warping.
Hence, $\omega$ must be tuned for the task at hand, ideally using expert knowledge.
Without the latter, one has to fallback on some automated approach.
In this paper, we evaluate $\ADTW$ on a classification benchmark (Section~\ref{sec:exp})
due to its clearly defined objective evaluation measures.
Our automated parameter selection method has been designed to work in this context.
It remains an important topic for future investigation on how to best parameterize ADTW in other scenarios.

The penalty range $0\leq\omega\leq\infty$ is continuous,
and hence cannot be exhaustively assessed unless the objective function is convex with respect to $\omega$,
or there are other properties that can be exploited for efficient complete search.
Instead, we use a similar method to the defacto standard for parameterizing $\CDTW$ and $\WDTW$.
To this end we define a range from small to large penalties.
We select from this range the one that achieves the lowest error
when used as the distance measure for nearest neighbor classification,
as assessed through leave-one-out cross validation (LOOCV) on the training set.

This leads to a major issue with an additive penalty such as $\omega$: its scale.
A small penalty in a given context may be huge in another one,
as the impact of the penalty depends on the magnitude of the costs at each step along a warping path.
Hence, we must find a reasonable way to determine the scale of penalties.
We achieved this by defining a maximal penalty  $\omega'$,
and multiplying it by a ratio $0\leq{}r\leq{}1$~\eqref{eq:omega}.
Our penalty range is now $0\leq\omega\leq\omega'$.
This leads to two questions: how to define $\omega'$, and how to sample $r$.
\begin{equation}\label{eq:omega}
    \omega = \omega' * r
\end{equation}

Let us first consider two series $S$ and $T$.
As $\ADTW_\omega(S,T)\leq\SQED(S,T)$, taking $\omega=\SQED(S,T)$ does ensure that $\ADTW_\omega(S,T) = \SQED(S,T)$.
Indeed, taking one single \emph{warping step} costs as much as the full diagonal.
In turn, this ensures that we do not need to test any other value beyond $\omega'=\SQED(S,T)$.
However, we need to have a single penalty for all series from $\Dtrain$,
so we sample random pairs of series, and set $\omega'$ to their average $\SQED$,
$\omega'=\mean_{S,T\sim\Dtrain}\SQED(S,T)$.

We now have to find the best $r$ --- we use a detour to better explain it.
Intuitively,  $\omega'$ represents the cost of $\ell$ steps along the diagonal,
and we can define an \emph{average step cost} $\alpha=\frac{\omega'}{\ell}$.
In turn, we have
\begin{equation}\label{eq:alpha}
    \omega = \alpha \times (\ell \times r')
\end{equation}
where $0\leq{}r'\leq{}1$.
As $\alpha$ remains a large penalty,
we need to start sampling $r_{1}'$ at value below $\frac{1}{\ell}$.
Without expert knowledge, our \emph{best guess} is to start at several order of magnitudes $m$ lower.
Simplifying equation~\eqref{eq:alpha} back to equation~\eqref{eq:omega},
$r_1$ must be small enough to account for both $m$, and the magnitude of $\ell$.
Then, successive $r_i$ must gradually increase toward $1$.
If several ``best'' $r_i$ are found, we take the median value.

As leave-one-out cross validation is time consuming,
and to ensure a fair comparison with the methods for parameterizing $\CDTW$ and $\WDTW$,
we limit the search to a hundred values~\cite{lines2015,tanFastEEFastEnsembles2020}.
The formula $r_{i}=(\frac{i}{100})^5$ for $1\leq{}i\leq{100}$ fulfills our requirements,
covering a range from $1E-10$ to 1 and favoring smaller penalties,
which tend to be more useful than large ones,
as any sufficiently large penalty confines the warping path to the diagonal.

\section{\label{sec:exp}Experiments}
Due to its clearly defined objective evaluation measures,
we evaluate $\ADTW$ against other distances on a classification benchmark,
using the UCR archive~\cite{dauUCRTimeSeries2019} in its 128 dataset versions.
We retained 112 datasets after filtering out those with series of variable length,
containing missing data, or having only one training exemplar per class
(leading to spurious results under LOOCV).
The distances are used in nearest neighbor classifiers.
The archive provides default train and test splits.
For each dataset, the parameters $w$, $g$ and $\omega$ are learned on the training data $\Dtrain$
following the methods described Sections~\ref{sec:relwork} and \ref{sec:params}.
We report the accuracy obtained on the test split.
Our results are available online~\cite{demoapp}.

\begin{figure}
    \centering
    \includegraphics[width=\textwidth]{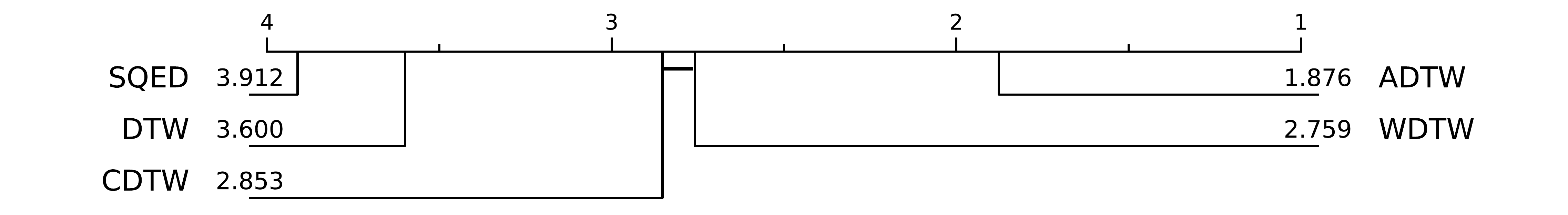}
    \caption{\label{fig:cd_d0}Test accuracy ranking of NN1 classifiers over 112 datasets from UCR128.}
\end{figure}

\begin{figure}
    \centering
    \includegraphics[width=\textwidth]{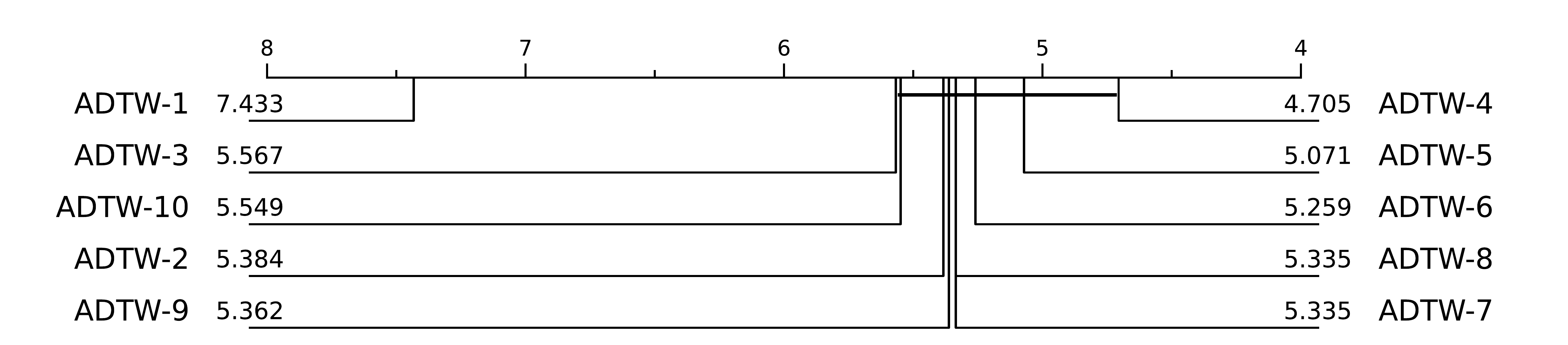}
    \caption{\label{fig:cd_adtw}
    Test accuracy ranking of $\ADTW$-$e$ when varying the exponent $e$ used at train time,
    over 112 datasets from UCR128
    }
\end{figure}

Following Demsar~\cite{Demsar2006}, we compare classifiers using the Wilcoxon signed-rank test.
The result can be graphically presented in mean-rank diagrams (Figures~\ref{fig:cd_d0} and~\ref{fig:cd_adtw}),
where classifiers not significantly different
(at a $0.05$ significance level, adjusted with the Holm correction for multiple testing)
are lined by a horizontal bar.
A rank closer to 1 indicates a more accurate classifier.

We compare $\ADTW$ against $\SQED$, $\DTW$, $\CDTW$ and $\WDTW$.
Figure~\ref{fig:cd_d0} summarizes the outcome of this comparison.
It shows that $\ADTW$ attains a significantly better rank on accuracy than any other measure.

\begin{figure}
    \captionsetup[subfigure]{aboveskip=-2pt,belowskip=7pt}
    \centering
    \begin{subfigure}[b]{0.49\textwidth}
        \includegraphics[width=\textwidth]{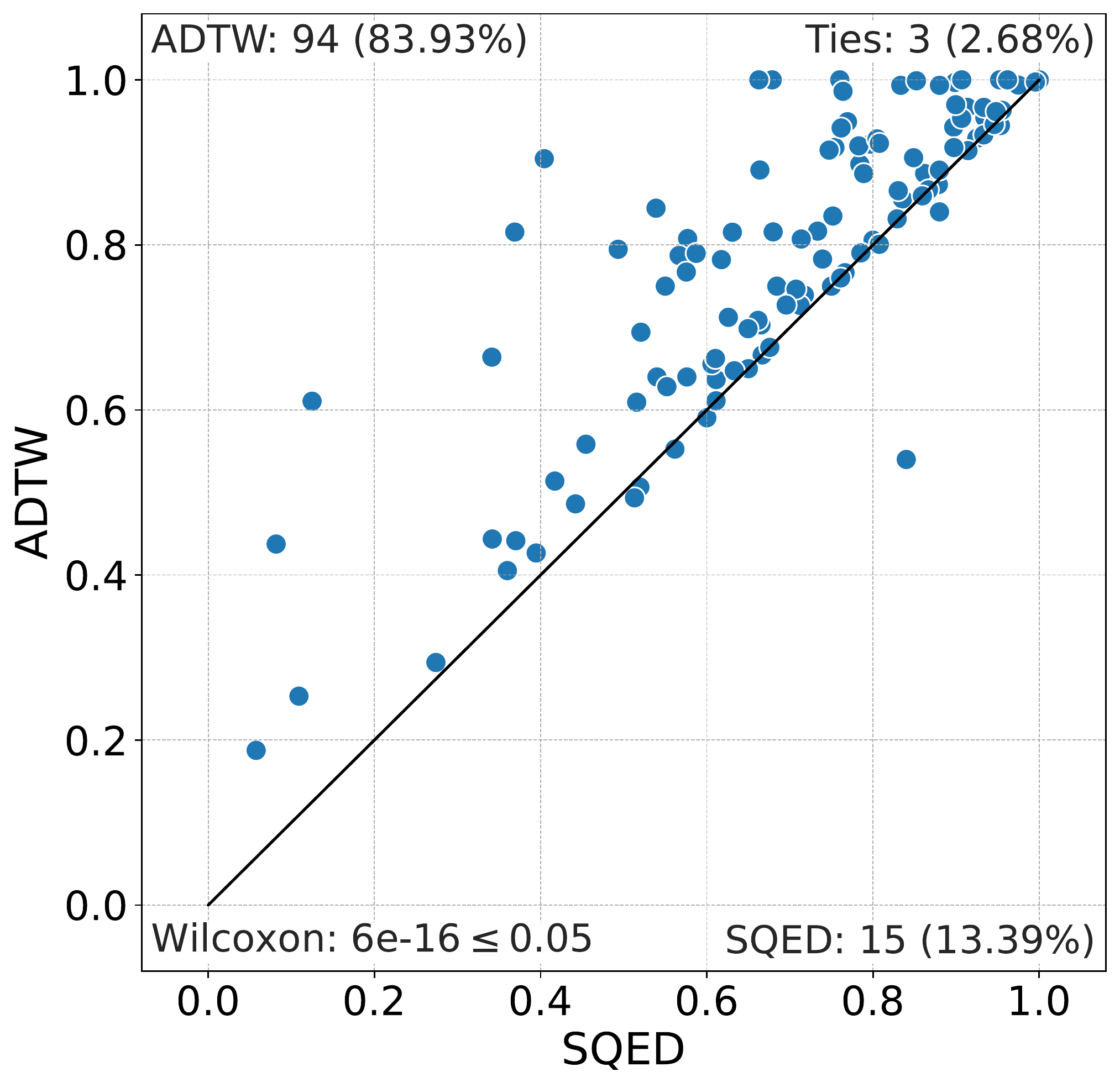}
        \caption{\label{fig:scp:sqed}$\ADTW$ vs. $\SQED$}
    \end{subfigure}
    \hfill
    \begin{subfigure}[b]{0.49\textwidth}
        \includegraphics[width=\textwidth]{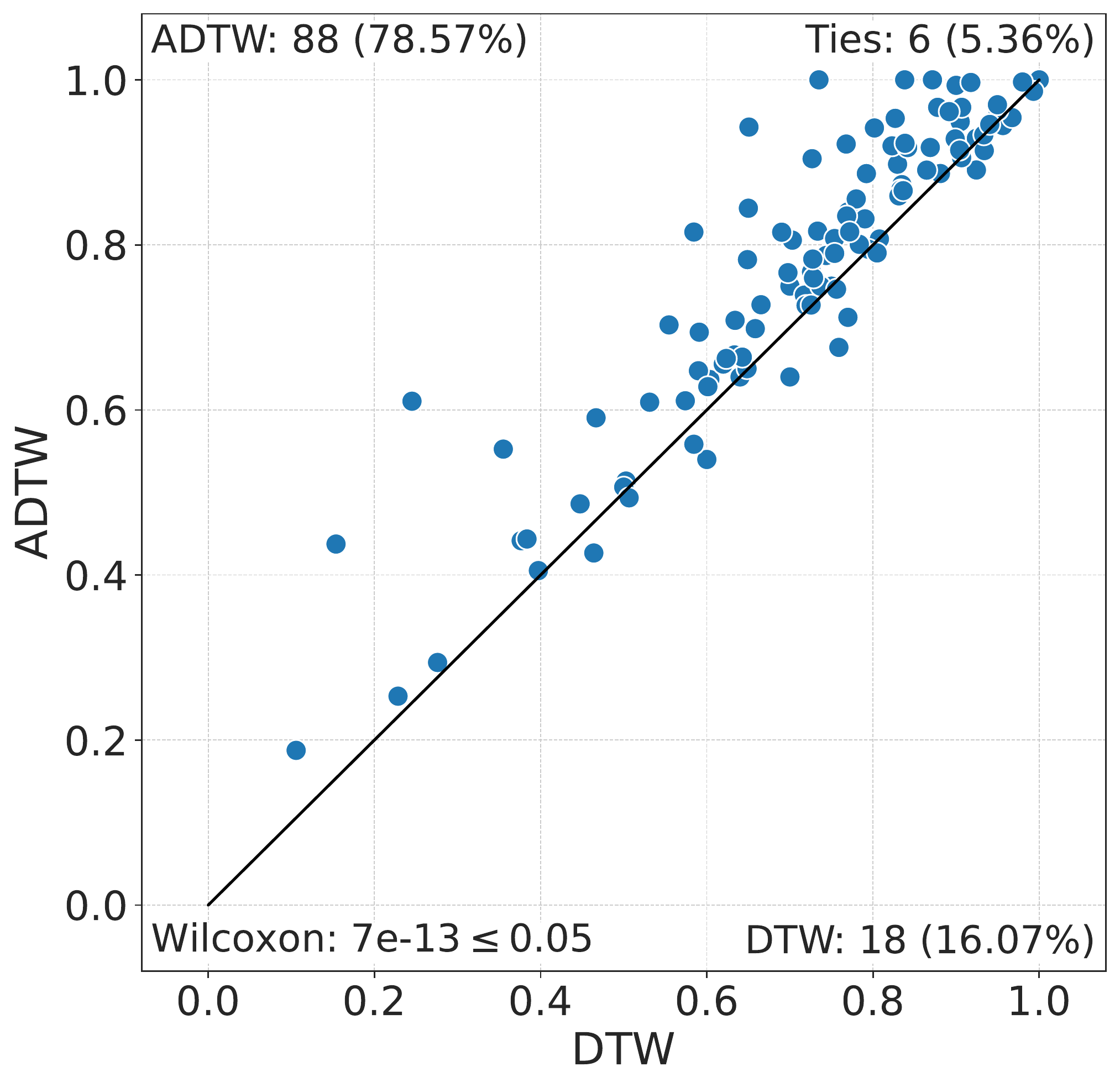}
        \caption{\label{fig:scp:dtw}$\ADTW$ vs. $\DTW$}
    \end{subfigure}
    \\
    \begin{subfigure}[b]{0.49\textwidth}
        \includegraphics[width=\textwidth]{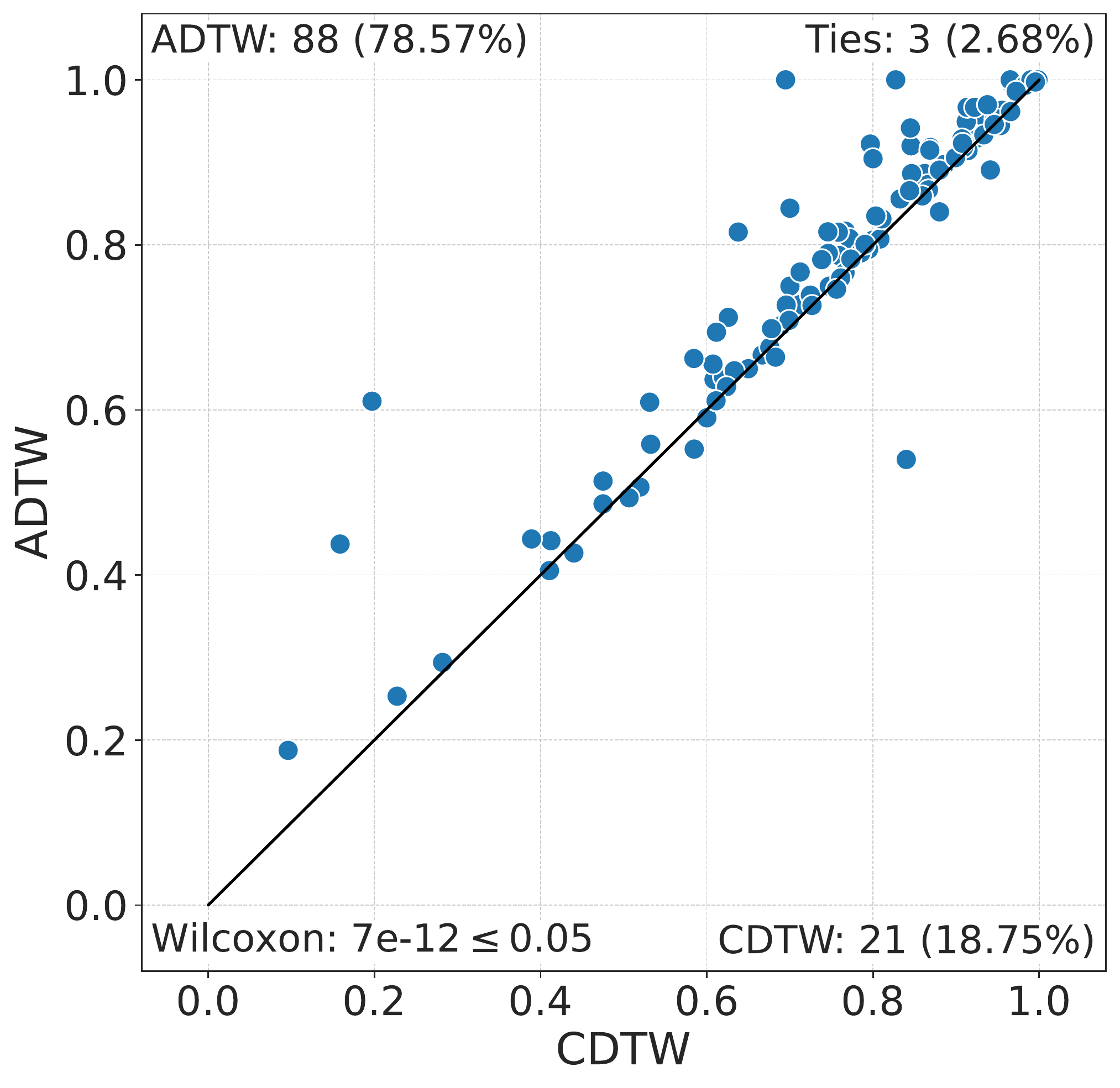}
        \caption{\label{fig:scp:cdtw}$\ADTW$ vs. $\CDTW$}
    \end{subfigure}
    \hfill
    \begin{subfigure}[b]{0.49\textwidth}
        \includegraphics[width=\textwidth]{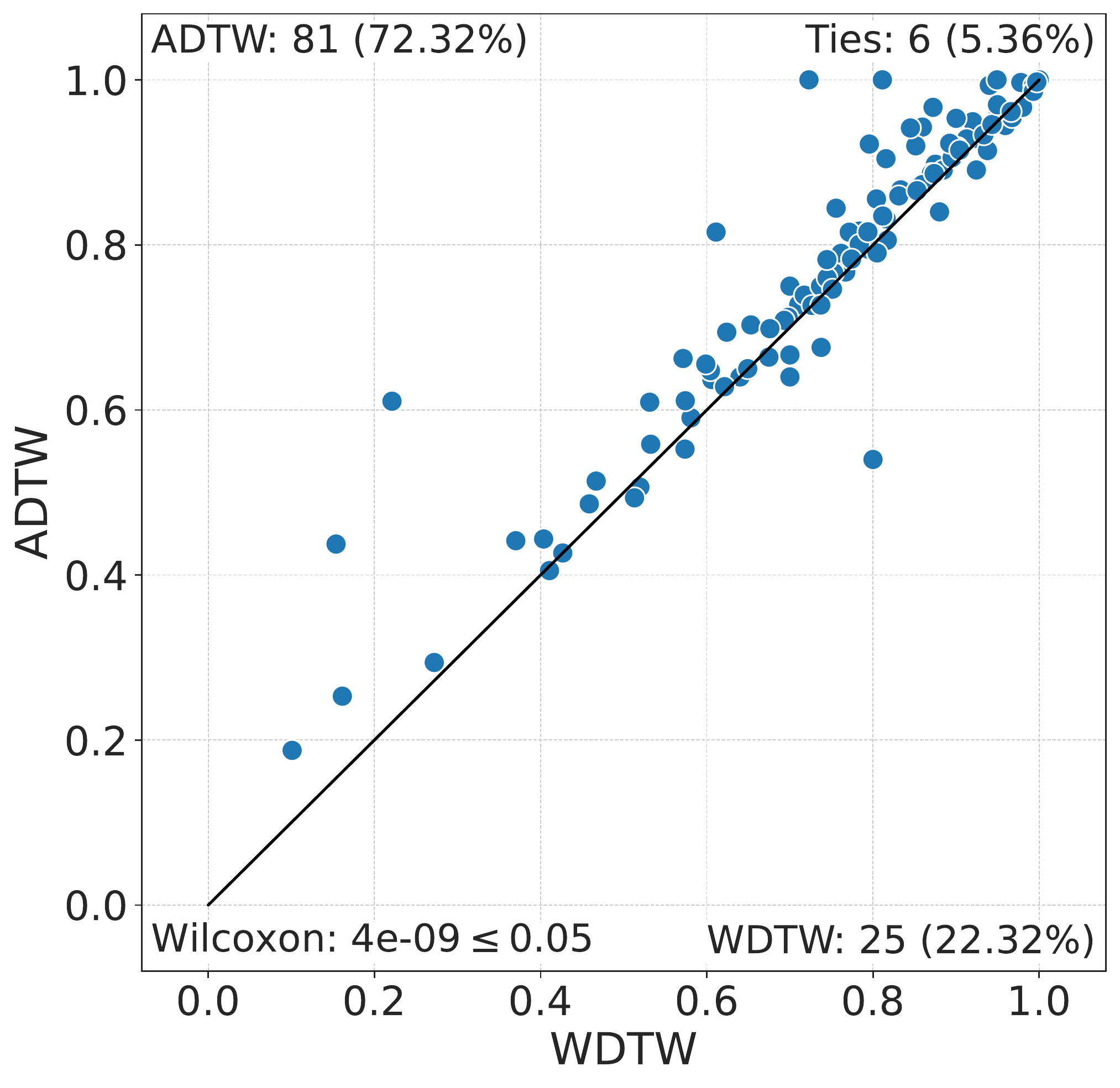}
        \caption{\label{fig:scp:wdtw}$\ADTW$ vs. $\WDTW$}
    \end{subfigure}
\vspace*{-10pt}
\caption{\label{fig:scp}
    Accuracy of NN1-$\ADTW$ vs.~NN1 with other distances on 112 datasets from UCR128.
    Each point represents a dataset.
    Points above the diagonal indicate that $\ADTW$
    gives better accuracy than the alternative, and reciprocally below the diagonal.
    }
\end{figure}

\begin{figure}
    \centering
    \begin{subfigure}[b]{0.49\textwidth}
        \includegraphics[width=\textwidth]{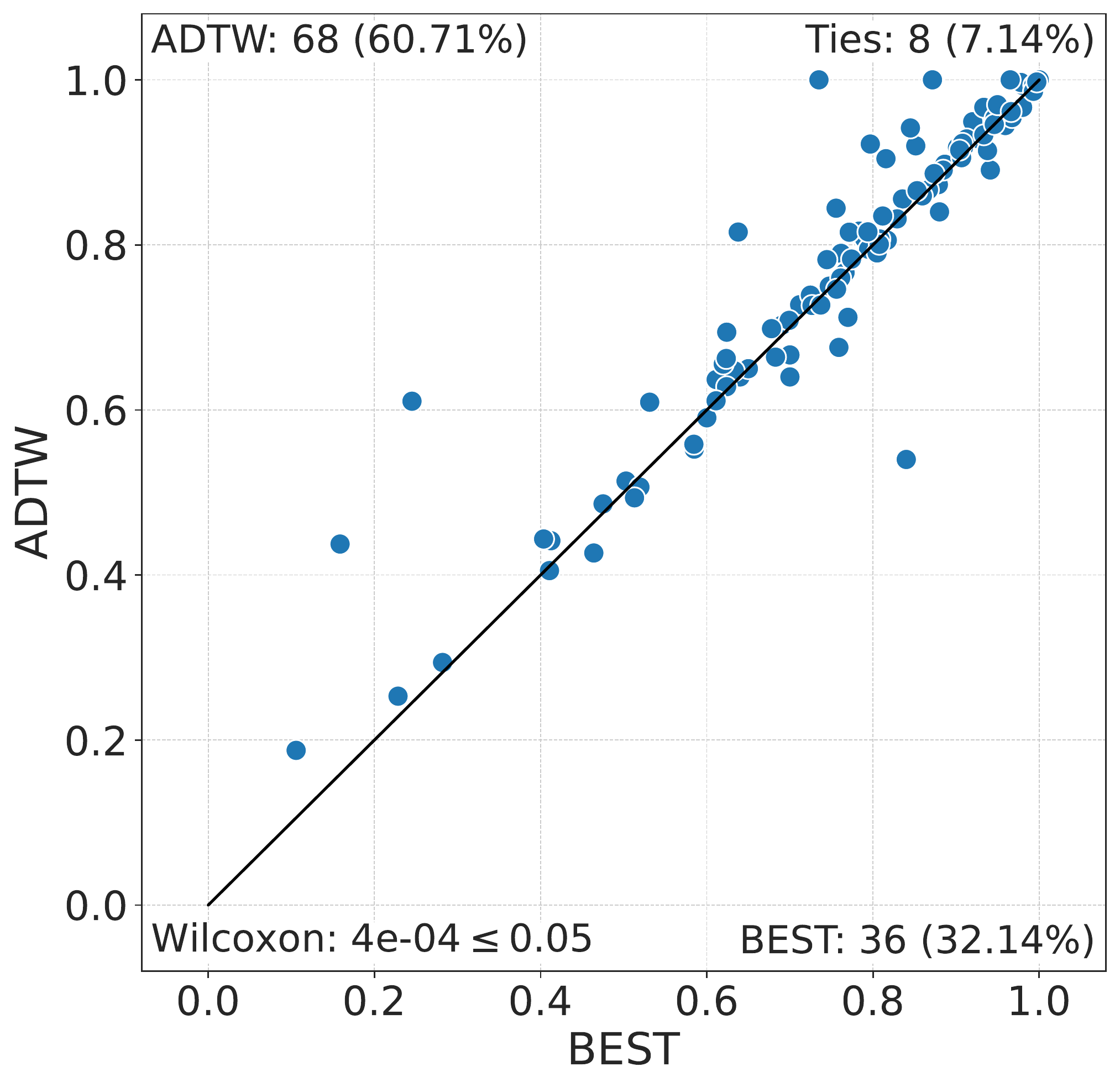}
        \caption{\label{fig:scp:bestd0}$\ADTW$ vs. $\BEST$}
    \end{subfigure}
    \caption{\label{fig:scp_best}
    Accuracy of NN1-$\ADTW$
    vs.~the best classifier per dataset among NN1-$\{\SQED, \DTW, \CDTW, \WDTW\}$, on 112 datasets from UCR128.
    A point above the diagonal indicates that $\ADTW$
    gives better accuracy than any alternative, and a point below the diagonal indicates that the best alternative gives higher accuracy than $\ADTW$.
    }
\end{figure}

Mean rank diagrams are useful to summarize relative performance,
but do not tell the full story
--- a classifier consistently outperforming another
would be deemed as significantly better ranked, even if the actual gain is minimal.
Accuracy scatter plots (see figures~\ref{fig:scp} and~\ref{fig:scp_best}) allow to visualize how
a classifier performs relatively to another one.
The space is divided by a diagonal representing equal accuracy, and each dot represents a dataset.
Points close to the diagonal indicates comparable accuracy between classifiers;
conversely, points far away indicates different accuracies.

Figure~\ref{fig:scp} shows the accuracy scatter plot of $\ADTW$ against other distances.
Points above the diagonal indicate datasets where $\ADTW$ is more accurate, and conversely below it.
We also indicate the numbers of ties and \emph{wins} per classifiers, and the resulting Wilcoxon score.
$\ADTW$ is almost always more accurate than $\SQED$ and $\DTW$ --- usually substantially so,
and the majority of points remain well above the diagonal for $\CDTW$ and $\WDTW$,
albeit by a smaller margin.
We note a point well below the diagonal in Figures~\ref{fig:scp:sqed} and~\ref{fig:scp:cdtw}.
This it the ``Rock'' dataset, for which our parameterization process fails to select a high enough penalty.
Although $\ADTW$ has the capacity to behave like $\SQED$ if appropriately parameterized, our parameterization process fails to achieve this for ``Rock.'' Effective parameterization remains an open problem.

We also compare against a hypothetical $\BEST$ classifier,
a classifier that uses the most accurate among all $\ADTW$ competitors.
Such a classifier is not feasible to obtain in practice
as it is not possible to be certain which classifier will obtain the lowest error on previously unseen test data.
The accuracy scatter plots against the hypothetical $\BEST$ classifier are shown in Figure~\ref{fig:scp_best}.
A~Wilcoxon signed-rank test assesses $\ADTW$ as significantly more accurate at a $0.05$ significance level 
than $\BEST$ ($p=4e^{-4}<0.05$).
These results suggest that $\ADTW$ is a good default choice of $\DTW$ variant.

Finally, we present a sensitivity analysis for the exponent used to tune $\omega$.
Figure~\ref{fig:cd_adtw} shows the effect of using different values for exponent $e$ when sampling
the ratio $r$ under LOOCV at train time.
All exponents above $1$ lead to comparable results. While $e=4$ leads to the best results on the benchmark, we have no theoretical grounds on which to prefer it and in other applications we have examined, slightly higher values lead to slightly greater accuracy.

\section{\label{sec:conclusion}Conclusions}
$\DTW$ is a popular time series distance measure that allows flexibility in the series alignments 
\cite{sakoe1971, sakoe1978}.
However, it can be too flexible for some applications,
and two main forms of constraint have been developed to limit the alignments.
Windows provide a strict limit on the distance over which points may be aligned
\cite{sakoe1978,itakuraMinimumPredictionResidual1975}.
Multiplicative weights penalize all off diagonal points proportionally to their distance from the diagonal
and the cost of their alignment \cite{jeongWeightedDynamicTime2011a}.

However, windows introduce an abrupt discontinuity.
Within the window there is unconstrained flexibility and beyond it there is none.
Multiplicative weights allow large warping at little cost if the aligned points have low cost.
Further, they penalize the length of an off-diagonal path rather than
the number of times the path deviates from the diagonal.
Whats more, neither windows nor multiplicative weights are symmetric with respect to
reversing the series
(they do not guarantee $\similarity(S,T)=\similarity(\reverse(S),\reverse(T))$ when $\ell_S\neq\ell_T$).

Amerced DTW introduces a tunable additive weight $\omega$
that produces a smooth range of distance measures such that
$\ADTW_\omega(S,T)$ is monotonic with respect to $\omega$,
$\ADTW_0(S,T)=\DTW(S,T)$,
and $\ADTW_{\infty}(S,T)=\SQED(S,T)$.
It is symmetric with respect to the order of the parameters and reversing the series,
irrespective of whether the series share the same length.
It has the intuitive property of penalizing the number of times a path deviates from simply aligning successive points in each series,
rather than penalizing the length of the paths following such a deviation.

Our experiments indicate that when $\ADTW$ is employed for nearest neighbor classification
on the widely used UCR benchmark,
it results in more accurate classification significantly more often than any other $\DTW$ variant.

The application of $\ADTW$ in the many other types of task to which $\DTW$
is often applied remains a productive direction for future investigation.
These include
similarity search \cite{rakthanmanon2012searching},
regression \cite{tan2021regression},
clustering \cite{petitjean2011global},
anomaly and outlier detection \cite{diab2019anomaly},
motif discovery \cite{alaee2021time}, 
forecasting \cite{bandara2021improving},
and subspace projection \cite{DENG2020107210}.
One issue that will need to be addressed in each of these domains is how best to tune the amercing penalty $\omega$,
especially if a task does not have objective criteria by which utility may be judged.
We hope that $\ADTW$ will prove as effective in these other applications
as it has proved to be in classification.
A C++ implementation of $\ADTW$ is available at \cite{demoapp}.

\section*{Acknowledgments}
\sloppy This work was supported by the Australian Research Council award DP210100072.
We would like to thank the maintainers and contributors for providing the UCR Archive,
and Hassan Fawaz for the critical diagram drawing code~\cite{cddiag}.
We are also grateful to Eamonn Keogh, Chang Wei Tan, and Mahsa Salehi for insightful comments on an early draft.

\bibliographystyle{elsarticle-num}
\bibliography{mybibfile}

\begin{thebibliography}{10}
\expandafter\ifx\csname url\endcsname\relax
  \def\url#1{\texttt{#1}}\fi
\expandafter\ifx\csname urlprefix\endcsname\relax\def\urlprefix{URL }\fi
\expandafter\ifx\csname href\endcsname\relax
  \def\href#1#2{#2} \def\path#1{#1}\fi

\bibitem{demoapp}
M.~Herrmann, Nn1 adtw demonstration application and results,
  \url{https://github.com/HerrmannM/paper-2021-ADTW} (2021).

\bibitem{sakoe1971}
H.~Sakoe, S.~Chiba, Recognition of continuously spoken words based on
  time-normalization by dynamic programming, Journal of the Acoustical Society
  of {Japan} 27~(9) (1971) 483--490.

\bibitem{sakoe1978}
H.~Sakoe, S.~Chiba, Dynamic programming algorithm optimization for spoken word
  recognition, IEEE Transactions on Acoustics, Speech, and Signal Processing
  26~(1) (1978) 43--49.
\newblock \href {https://doi.org/10.1109/TASSP.1978.1163055}
  {\path{doi:10.1109/TASSP.1978.1163055}}.

\bibitem{cheng2016image}
H.~Cheng, Z.~Dai, Z.~Liu, Y.~Zhao, An image-to-class dynamic time warping
  approach for both 3d static and trajectory hand gesture recognition, Pattern
  Recognition 55 (2016) 137--147.

\bibitem{OKAWA2021107699}
M.~Okawa, Online signature verification using single-template matching with
  time-series averaging and gradient boosting, Pattern Recognition 112 (2021)
  107699.

\bibitem{yasseen2016shape}
Z.~Yasseen, A.~Verroust-Blondet, A.~Nasri, Shape matching by part alignment
  using extended chordal axis transform, Pattern Recognition 57 (2016)
  115--135.

\bibitem{singh2017smart}
G.~Singh, D.~Bansal, S.~Sofat, N.~Aggarwal, Smart patrolling: An efficient road
  surface monitoring using smartphone sensors and crowdsourcing, Pervasive and
  Mobile Computing 40 (2017) 71--88.

\bibitem{cao2016real}
Y.~Cao, N.~Rakhilin, P.~H. Gordon, X.~Shen, E.~C. Kan, A real-time spike
  classification method based on dynamic time warping for extracellular enteric
  neural recording with large waveform variability, Journal of Neuroscience
  Methods 261 (2016) 97--109.

\bibitem{varatharajan2018wearable}
R.~Varatharajan, G.~Manogaran, M.~K. Priyan, R.~Sundarasekar, Wearable sensor
  devices for early detection of alzheimer disease using dynamic time warping
  algorithm, Cluster Computing 21~(1) (2018) 681--690.

\bibitem{jeongWeightedDynamicTime2011a}
Y.-S. Jeong, M.~K. Jeong, O.~A. Omitaomu, Weighted dynamic time warping for
  time series classification, Pattern Recognition 44~(9) (2011) 2231--2240.
\newblock \href {https://doi.org/10.1016/j.patcog.2010.09.022}
  {\path{doi:10.1016/j.patcog.2010.09.022}}.

\bibitem{rakthanmanon2012searching}
T.~Rakthanmanon, B.~Campana, A.~Mueen, G.~Batista, B.~Westover, Q.~Zhu,
  J.~Zakaria, E.~Keogh, Searching and mining trillions of time series
  subsequences under dynamic time warping, in: Proc.\ 18th ACM SIGKDD Int.\
  Conf.\ Knowledge Discovery and Data Mining, 2012, pp. 262--270.

\bibitem{tan2021regression}
C.~W. Tan, C.~Bergmeir, F.~Petitjean, G.~I. Webb, Time series extrinsic
  regression, Data Mining and Knowledge Discovery 35~(3) (2021) 1032--1060.
\newblock \href {https://doi.org/10.1007/s10618-021-00745-9}
  {\path{doi:10.1007/s10618-021-00745-9}}.

\bibitem{petitjean2011global}
F.~Petitjean, A.~Ketterlin, P.~Gan{\c{c}}arski, A global averaging method for
  dynamic time warping, with applications to clustering, Pattern Recognition
  44~(3) (2011) 678--693.

\bibitem{diab2019anomaly}
D.~M. Diab, B.~AsSadhan, H.~Binsalleeh, S.~Lambotharan, K.~G. Kyriakopoulos,
  I.~Ghafir, Anomaly detection using dynamic time warping, in: 2019 IEEE
  International Conference on Computational Science and Engineering (CSE) and
  IEEE International Conference on Embedded and Ubiquitous Computing (EUC),
  IEEE, 2019, pp. 193--198.

\bibitem{alaee2021time}
S.~Alaee, R.~Mercer, K.~Kamgar, E.~Keogh, Time series motifs discovery under
  {DTW} allows more robust discovery of conserved structure, Data Mining and
  Knowledge Discovery 35~(3) (2021) 863--910.

\bibitem{bandara2021improving}
K.~Bandara, H.~Hewamalage, Y.-H. Liu, Y.~Kang, C.~Bergmeir, Improving the
  accuracy of global forecasting models using time series data augmentation,
  Pattern Recognition 120 (2021) 108148.

\bibitem{DENG2020107210}
H.~Deng, W.~Chen, Q.~Shen, A.~J. Ma, P.~C. Yuen, G.~Feng, Invariant subspace
  learning for time series data based on dynamic time warping distance, Pattern
  Recognition 102 (2020) 107210.
\newblock \href {https://doi.org/https://doi.org/10.1016/j.patcog.2020.107210}
  {\path{doi:https://doi.org/10.1016/j.patcog.2020.107210}}.

\bibitem{herrmann2021early}
M.~Herrmann, G.~I. Webb,
  \href{https://doi.org/10.1007/s10618-021-00782-4}{Early abandoning and
  pruning for elastic distances including dynamic time warping}, Data Mining
  and Knowledge Discovery 35~(6) (2021) 2577--2601.
\newblock \href {https://doi.org/10.1007/s10618-021-00782-4}
  {\path{doi:10.1007/s10618-021-00782-4}}.
\newline\urlprefix\url{https://doi.org/10.1007/s10618-021-00782-4}

\bibitem{itakuraMinimumPredictionResidual1975}
F.~Itakura, Minimum prediction residual principle applied to speech
  recognition, IEEE Transactions on Acoustics, Speech, and Signal Processing
  23~(1) (1975) 67--72.
\newblock \href {https://doi.org/10.1109/TASSP.1975.1162641}
  {\path{doi:10.1109/TASSP.1975.1162641}}.

\bibitem{lines2015}
J.~Lines, A.~Bagnall, Time series classification with ensembles of elastic
  distance measures, Data Mining and Knowledge Discovery 29~(3) (2015)
  565--592.
\newblock \href {https://doi.org/10.1007/s10618-014-0361-2}
  {\path{doi:10.1007/s10618-014-0361-2}}.

\bibitem{lucasProximityForestEffective2019}
B.~Lucas, A.~Shifaz, C.~Pelletier, L.~O'Neill, N.~Zaidi, B.~Goethals,
  F.~Petitjean, G.~I. Webb, Proximity {{Forest}}: An effective and scalable
  distance-based classifier for time series, Data Mining and Knowledge
  Discovery 33~(3) (2019) 607--635.
\newblock \href {https://doi.org/10.1007/s10618-019-00617-3}
  {\path{doi:10.1007/s10618-019-00617-3}}.

\bibitem{shifazTSCHIEFScalableAccurate2020}
A.~Shifaz, C.~Pelletier, F.~Petitjean, G.~I. Webb, {TS-CHIEF}: A scalable and
  accurate forest algorithm for time series classification, Data Mining and
  Knowledge Discovery 34~(3) (2020) 742--775.
\newblock \href {https://doi.org/10.1007/s10618-020-00679-8}
  {\path{doi:10.1007/s10618-020-00679-8}}.

\bibitem{dauUCRTimeSeries2019}
H.~A. Dau, A.~Bagnall, K.~Kamgar, C.-C.~M. Yeh, Y.~Zhu, S.~Gharghabi, C.~A.
  Ratanamahatana, E.~Keogh, The {{UCR Time Series Archive}}, arXiv:1810.07758
  [cs, stat] (Sep. 2019).
\newblock \href {http://arxiv.org/abs/1810.07758} {\path{arXiv:1810.07758}}.

\bibitem{tanFastEEFastEnsembles2020}
C.~W. Tan, F.~Petitjean, G.~I. Webb, {{FastEE}}: Fast {{Ensembles}} of
  {{Elastic Distances}} for time series classification, Data Mining and
  Knowledge Discovery 34~(1) (2020) 231--272.
\newblock \href {https://doi.org/10.1007/s10618-019-00663-x}
  {\path{doi:10.1007/s10618-019-00663-x}}.

\bibitem{Demsar2006}
J.~Dem{\v {s}}ar, {Statistical Comparisons of Classifiers over Multiple Data
  Sets}, Journal of Machine Learning Research 7 (2006) 1--30.

\bibitem{cddiag}
H.~Ismail~Fawaz, Critical difference diagram with {Wilcoxon-Holm} post-hoc
  analysis, \url{https://github.com/hfawaz/cd-diagram} (2019).

\end{thebibliography}
\end{document}